\documentclass[final,5p,times,twocolumn]{elsarticle}
\usepackage{amsmath,amssymb,amsfonts}
\usepackage{graphicx}
\usepackage{booktabs}
\usepackage{algorithm}
\usepackage{caption}
\usepackage{subcaption}
\usepackage[hidelinks]{hyperref}
\usepackage{algorithmic}
\usepackage{enumitem} 
\usepackage{float}
\journal{Neurocomputing}

\begin{document}

\begin{frontmatter}

\title{Synergizing Kolmogorov-Arnold Networks with Dynamic Adaptive Weighting for High-Frequency and Multi-Scale PDE Solutions}

\author[1]{Guokan Chen}
\ead{cgk200409@163.com}

\author[1]{Yao Xiao\corref{cor1}}
\ead{yaox@fjut.edu.cn}

\author[1]{Bin Fan}
\ead{bfan@fjut.edu.cn}

\author[1]{Meixin Xiong}
\ead{xmx2023@fjut.edu.cn}

\author[1]{Zhicheng Lin}
\ead{flyriverms@qq.com}

\author[1]{Yuanying Liu}
\ead{1143104558@qq.com}

\cortext[cor1]{Corresponding author}
\address[1]{School of Computer Science and Mathematics, Fujian University of Technology, Fuzhou, 350118, China}
\begin{abstract}
PINNs enhance scientific computing by incorporating physical laws into neural network structures, leading to significant advancements in scientific computing. However, PINNs struggle with multi-scale and high-frequency problems due to pathological gradient flow and spectral bias, which severely limit their predictive power. By combining an enhanced network architecture with a dynamically adaptive weighting mechanism featuring upper-bound constraints, we propose the Dynamic Balancing Adaptive Weighting Physics-Informed Kolmogorov-Arnold Network (DBAW-PIKAN). The proposed method effectively mitigates gradient-related failure modes and overcomes bottlenecks in function representation. Compared to baseline models, the proposed method accelerates the convergence process and improves solution accuracy by at least an order of magnitude without introducing additional computational complexity. Numerical results on the Klein-Gordon, Burgers, and Helmholtz equations demonstrate that DBAW-PIKAN achieves superior accuracy and generalization performance.
\end{abstract}


\end{frontmatter}


\section{Introduction}
\label{sec1}
Partial differential equations (PDEs) describe the continuous evolution of physical quantities in space and time through the relationship between unknown functions and their partial derivatives\cite{ren2025physics}. The practical solution of PDEs is severely constrained by in computational efficiency and geometric adaptability with traditional mesh-based methods. Physics-Informed Neural Networks (PINNs) integrate physical constraints into the loss function, enabling solutions that possess both physical interpretability and numerical advantages.  This method alleviates the challenges faced by traditional mesh-based methods in high-dimensional modeling, complex boundary handling, and parameter inversion, notably their high computational costs, difficulties in mesh adaptation, and stability problems, thereby directly addressing these bottlenecks\cite{cuomo2022scientific}. Since the inception of the PINN framework, it has demonstrated remarkable adaptability and potential across various multidisciplinary scientific and engineering applications\cite{karniadakis2021physics},including fluid dynamics\cite{raissi2019deep}, biomedical engineering, and geosciences. For instance, the Karniadakis group applied PINNs to the numerical simulation of complex fluid dynamics, successfully achieving efficient solutions for highly nonlinear problems such as unsteady flows and turbulence \cite{jin2021nsfnets}. In the field of biomedical engineering, PINNs have been employed for simulating hemodynamic evolution and analyzing the mechanical properties of soft tissues, providing new technical support for clinical diagnosis and medical device design \cite{dousse2022multi, peng2020multiscale}. Additionally, this method has been successfully applied to dynamic process simulation in geosciences \cite{he2021physics}, structural stress analysis in solid mechanics \cite{haghighat2021physics}, and multi-scale heat conduction problems \cite{cai2021flow}, fully validating its cross-disciplinary universality and engineering practicality.

Although widely applied, the performance of PINNs in solving multi-scale or strongly nonlinear PDE problems is constrained by two primary factors. First, Wang et al.\cite{wang2021understanding} pointed out that PINNs are subject to gradient flow pathologies during training\cite{wang2022and}. This phenomenon stems from the imbalance in gradient magnitudes across different terms of the loss function. Gradient magnitudes across components of a composite loss function can differ by several orders of magnitude, leading to gradient dominance and optimization Therefore, dynamically coordinating the contribution weights of various loss terms during training has become a prerequisite for the successful application of PINNs. Moreover, the spectral bias existing in multilayer perceptrons limits the capability of PINNs \cite{rahaman2019spectral, cao2019towards}. This leads to a tendency to capture low-frequency macroscopic trends, while failing to resolve high-frequency details and local discontinuities in solutions to multi-scale PDEs or problems with steep gradients. To mitigate this architectural bottleneck, Kolmogorov-Arnold Networks (KANs) have emerged as a powerful alternative. Specifically, recent advancements such as Wavelet KAN (WAV-KAN) \cite{seydi2024unveiling} integrate wavelet basis functions into the KAN framework, providing inherent multi-resolution analysis capabilities. Unlike the global activation functions in MLPs, the localized nature of wavelets allows WAV-KAN to capture high-frequency oscillations and sharp transitions naturally from an architectural perspective, effectively bypassing the spectral bias inherent in traditional neural architectures. Consequently, the computational efficiency and solution accuracy of PINNs are affected for such problems. The performance issue is further compounded by the computational mechanism of PINNs, which rely on automatic differentiation (AD) to calculate PDE residuals. The higher-order differential operations involved in AD can amplify errors stemming from the network's inadequate representation of high-frequency components. This amplification leads to a significant discrepancy between the estimated and true residual values, which in turn compromises the optimizer and disrupts the overall training dynamics. Consequently, this intrinsic mechanism fundamentally limits the convergence rate and final approximation accuracy of MLP-based PINNs.

To address the problems mentioned above, existing studies have adopted two main approaches: optimizing training dynamics and improving the representational capacity of network architectures. Regarding the optimization of training dynamics to mitigate gradient flow pathologies, various dynamic adaptive weighting strategies have been developed to balance the contributions of different loss terms during optimization. These approaches often draw inspiration from multi-task learning, such as employing uncertainty-based weighting \cite{chen2025self} or designing heuristic algorithms for self-adaptive weight adjustment \cite{mcclenny2023selfadaptive, xiang2022selfadaptive}. In terms of enhancing architectural expressivity, one prevalent approach involves improving input representations, such as introducing Fourier feature encoding \cite{tancik2020fourier,wang2021eigenvector} to facilitate the learning of high-frequency functions.  However, this method does not fundamentally resolve the spectral bias of neural networks; rather, it shifts or complexifies the issue. In some cases, this can lead to training dynamics that are difficult to predict and control. Therefore, researchers are committed to pursuing fundamental solutions by innovating at the foundational level of network architecture.  Recently, Kolmogorov-Arnold Networks (KAN), proposed by Liu et al. \cite{liu2024kan} represent a significant breakthrough in this direction. Unlike MLPs, KAN places learnable spline-based activation functions on the "edges" of the network rather than using fixed activations on the "nodes." This design exhibits significant potential in terms of function approximation accuracy and parameter efficiency \cite{wang2025kolmogorov,shuai2025physics}, offering a viable path to overcoming the expressivity bottlenecks of MLPs.

Adaptive weighting strategies and KAN networks primarily address the issues of pathological gradient flow and spectral bias, respectively, yet neither approach is capable of simultaneously resolving both problems. The potential of an expressive architecture, such as KAN, cannot be fully realized if its training process remains unstable. Conversely, an effective weighting strategy applied to an architecture with limited expressivity, such as an MLP, will inevitably be constrained by a lower performance ceiling. Therefore, we propose an integrated Dynamic Balancing Adaptive Weighting Physics-Informed Kolmogorov-Arnold Network (DBAW-PIKAN) framework, aimed at simultaneously addressing the dual challenges of architectural expressivity limitations and optimization imbalance inherent in PINNs. The core of the proposed framework lies in the adoption of KAN \cite{liu2024kan} as a high-expressivity function approximator to overcome spectral bias, coupled with a novel DBAW strategy to stabilize the training process. DBAW constrains adaptive weights by introducing a dynamically decaying upper bound, effectively mitigating pathological gradient flow while preventing the weight explosion risk commonly associated with traditional methods.  Subsequent experimental results will demonstrate that the synergy between this architecture and the optimization strategy yields significant improvements in both solution accuracy and training stability compared to baseline models.

\section{Methodology}
We propose a DBAW-PIKAN framework to address the dual challenges of representational capacity and multi-objective optimization balance encountered by PINNs. A novel dynamic adaptive weighting strategy is introduced to optimize the training of the Physics-Informed Kolmogorov-Arnold Network (PIKAN) architecture, which constitutes the core of the proposed framework. 

In this section, we present a systematic overview of the proposed integrated framework. Section 2.1 first reviews the standard PINNs. Section 2.2 introduces the PIKAN framework in detail, which serves as the cornerstone of the model's expressive power. Section 2.3 elaborates on the DBAW strategy designed to stabilize the optimization process. Finally, Section 2.4 demonstrates how these components are synergistically combined to construct the ultimate DBAW-PIKAN framework.
\subsection{Physics-informed neural networks}
PINNs aim to infer a continuous latent function $u(x, t)$ that represents the solution to a system of partial differential equations (PDEs).  A general nonlinear PDE system can be expressed as:
\begin{equation}u_{t}+\mathcal{N}_{x}[u]=0, \quad x \in \Omega, t \in T,
\end{equation}
where $u_t$ denotes the partial derivative of $u$ with respect to time $t$, and $\mathcal{N}_x[\cdot]$ is a linear or nonlinear differential operator acting on the spatial variable $x$ and its derivatives. To obtain a well-posed solution, the system must satisfy a set of initial conditions and boundary conditions. The initial conditions are defined as:
\begin{equation}
u(x, t_0) = h(x), \quad x \in \Omega
\end{equation}and the  boundary conditions are denoted as:
\begin{equation}
u(x, t) = g(x, t), \quad x \in \partial\Omega, t \in T.
\end{equation}

During the training of PINNs, the solution $u(x, t)$ is approximated by a deep neural network $u_{\theta}(x, t)$, where $\theta$ represents the trainable parameters of the network. The core of this approach lies in embedding the physical laws, which including the governing PDEs, initial conditions, and boundary conditions, into the learning process of network as soft constraints. This is achieved by defining a composite loss function $\mathcal{L}(\theta)$, which typically consists of three components: the PDE residual loss ($\mathcal{L}_r$), the initial condition loss ($\mathcal{L}_{ic}$), and the boundary condition loss ($\mathcal{L}_{bc}$). $\mathcal{L}_r$ penalizes the extent to which the neural network $u_{\theta}$ violates the governing PDE within the domain, while $\mathcal{L}_{ic}$ and $\mathcal{L}_{bc}$ penalize the discrepancies between $u_{\theta}$ and the specified initial and boundary conditions, respectively. These loss terms are commonly defined using the Mean Squared Error (MSE) \cite{Raissi2017PINNpart1}:
\begin{equation}
\mathcal{L}_{r}=\frac{1}{N_{r}}\sum_{i=1}^{N_{r}}|r_{\theta}(x_{r}^{i},t_{r}^{i})|^{2},
\end{equation}
\begin{equation}
\mathcal{L}_{bc}=\frac{1}{N_{bc}}\sum_{i=1}^{N_{bc}}|u_{\theta}(x_{bc}^{i},t_{b}^{i})-g(x_{b}^{i},t_{b}^{i})|^{2},
\end{equation}
\begin{equation}
\mathcal{L}_{ic}=\frac{1}{N_{ic}}\sum_{i=1}^{N_{ic}}|u_{\theta}(x_{ic}^{i},t_0)-h(x_{ic}^{i})|^{2},
\end{equation}
where $N_r$, $N_{ic}$, and $N_{bc}$ denote the total number of sampling points for each respective loss term. The sets $\{(x_{r}^{i},t_{r}^{i})\}_{i=1}^{N_{r}}$, $\{(x_{ic}^{i})\}_{i=1}^{N_{ic}}$, and $\{(x_{bc}^{i},t_{b}^{i})\}_{i=1}^{N_{bc}}$ represent the collocation points within the domain $\Omega \times T$, the data points at the initial time $t=t_0$, and the data points on the boundary $\partial\Omega \times T$, respectively. The PDE residual $r_{\theta}(x,t)$ is defined as:
\begin{equation}
r_{\theta}(x,t):=\frac{\partial}{\partial t}u_{\theta}(x,t)+\mathcal{N}_{x}[u_{\theta}(x,t)].
\end{equation}

The partial derivatives of the network output $u_{\theta}$ with respect to its spatio-temporal input coordinates (e.g., $\frac{\partial u_{\theta}}{\partial t}$) can be accurately computed using automatic differentiation (AD) techniques available in modern deep learning frameworks.Ultimately, the total loss function $\mathcal{L}(\theta)$ is defined as the weighted sum of the aforementioned components:
\begin{equation}
\mathcal{L}(\theta) = w_{r}\mathcal{L}_{r} + w_{ic}\mathcal{L}_{ic} + w_{bc}\mathcal{L}_{bc}.
\end{equation}

\begin{figure*}
    \centering
    \includegraphics[width=1\linewidth]{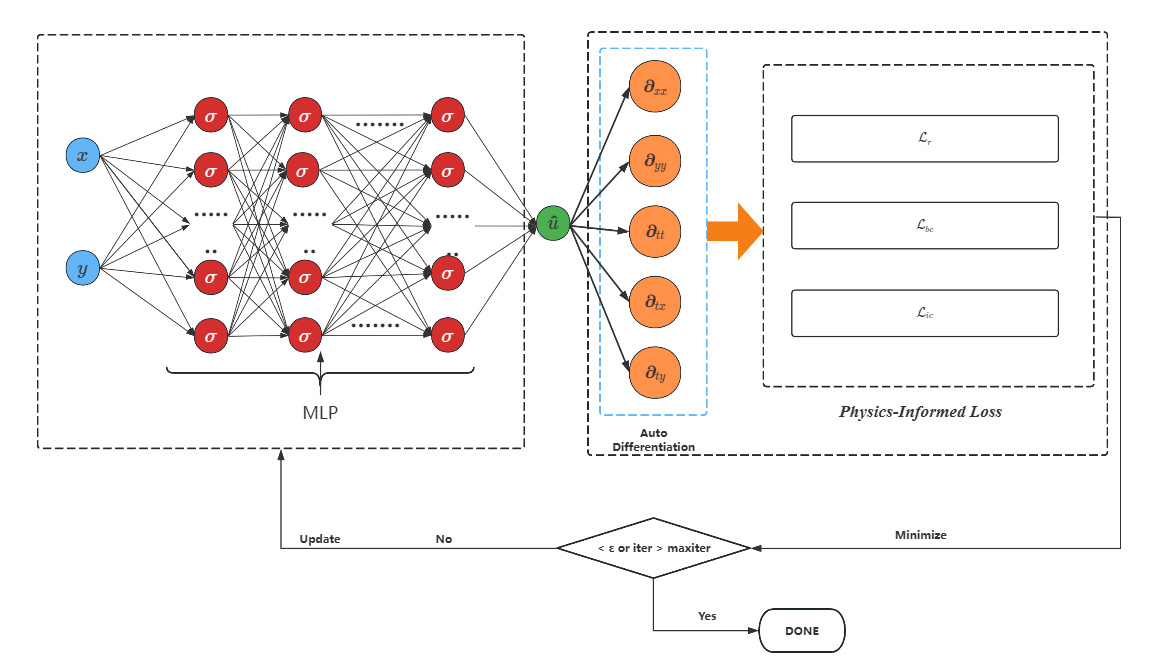}
    \caption{PINNs Framework Diagram}
    \label{fig_pinn_KUANGJIA}
\end{figure*}

The workflow of this framework is illustrated in Fig. \ref{fig_pinn_KUANGJIA}. The optimization objective for the network parameters $\theta$ is to minimize this total loss function, a process typically executed via gradient-based optimization algorithms such as the Adam optimizer. In the standard PINNs framework, the weights $w_r, w_{ic},$ and $w_{bc}$ are usually treated as fixed hyperparameters, the determination of which often relies on time-consuming and empirical manual tuning. As previously noted, the significant discrepancies in magnitude and convergence rates among different loss terms constitute one of the critical challenges leading to training instability and hindering model convergence.
\subsection{Physics-Informed Kolmogorov-Arnold Network}
To address the inherent limitations in the expressive power of traditional Multi-Layer Perceptrons, particularly the spectral bias problem where the network struggles to learn complex functions containing multi-scale or high-frequency information, we introduce Kolmogorov-Arnold Networks as a high-capacity function approximator \cite{rahaman2019spectral}. Unlike MLPs, which concentrate nonlinear transformations at nodes with fixed activation functions, KANs place learnable nonlinear activation functions on the edges of the network. The theoretical foundation of KAN is the Kolmogorov-Arnold Representation Theorem (KART)\cite{arnold1957}, which states that any multivariate continuous function $f(\mathbf{x})$ can be represented as a composition of a finite number of univariate functions and addition operations:
\begin{equation}
f(\mathbf{x}) = \sum_{q=1}^{2n+1} \Phi_q \left( \sum_{p=1}^{n} \phi_{q,p}(x_p) \right).
\end{equation}
A deep KAN is composed of multiple KAN layers $\Psi$, denoted as $\text{KAN}(\mathbf{x}) = (\Psi_L \circ \cdots \circ \Psi_1)(\mathbf{x})$. For a single KAN layer mapping an $n_{\text{in}}$-dimensional input $\mathbf{x}_{\text{in}}$ to an $n_{\text{out}}$-dimensional output, the mapping $\Psi$ is expressed as:
\begin{equation}
\psi_j(\mathbf{x}_{\text{in}}) = \sum^{i=1}{n_{\text{in}}} \phi_{j,i}(x_i), \quad j = 1, \ldots, n_{\text{out}},
\end{equation}
where $\phi_{j,i}(\cdot)$ are learnable univariate activation functions on the edges. In practice, these functions are parameterized as a linear combination of B-spline basis functions $B_{k}(x)$:
\begin{equation}
\phi(x) = w_b \cdot b(x) + \sum_{k=1}^{G} c_k \cdot B_{k}(x),
\end{equation}
where $b(x)$ is a fixed basis function (e.g., SiLU) and the coefficients $c_k$ of the B-spline part are trainable parameters. In the context of solving PDEs, B-splines are selected due to their advantages in numerical stability, local adaptability, and derivative computation efficiency\cite{toshniwal2020multi}. Firstly, B-splines possess the local support property, meaning a single basis function is non-zero only within a finite interval. This implies that modifying a coefficient $c_k$ only affects the function's shape in a local region. Since PDE solutions often feature localized sharp variations (such as shock waves or boundary layers), the locality of B-splines enables KAN to adaptively capture these phenomena with high parameter efficiency, thereby mitigating the spectral bias of MLPs. Additionally, regarding numerical stability, B-splines, as piecewise polynomials, effectively avoid the Runge phenomenon, which refers to severe oscillations near the endpoints of an interval and is common in high-order global polynomial interpolation.\cite{magueresse2025energy} .
In the PINNs framework, unstable function approximators adversely affect the gradients computed via automatic differentiation (AD); thus, the stability of B-splines is critical for ensuring a robust optimization process. Furthermore, calculating PDE residuals requires high-order derivatives of the network output. The derivatives of B-splines have closed-form analytical expressions, allowing any order of derivative to be computed directly, accurately, and efficiently. This provides superior performance and more robust gradient estimation compared to methods relying solely on AD \cite{beccari2022stable}.

Based on these advantages, the Physics-Informed Kolmogorov-Arnold Network (PIKAN) framework is constructed. To enhance training stability, we implement a specific hierarchical structure proposed by Wang et al., which introduces a residual connection and a tanh activation function after each KAN layer. The forward propagation of a single layer $\mathcal{H}^{(l)}$ is defined as:
\begin{equation}
\mathbf{X}^{(l+1)} = \tanh \Bigl( \sum \phi^{(l+1)}(\mathbf{X}^{(l)}) + \mathbf{W}^{(l+1)} \sigma(\mathbf{X}^{(l)}) \Bigr),
\end{equation}
where $W^{(l+1)}\cdot\sigma(X^{(l)})$ constitutes a ResNet-like residual term intended to alleviate vanishing gradient issues in deep networks. The tanh function constrains the output to the range $[-1, 1]$, ensuring that inputs to subsequent B-spline activations remain within the predefined effective grid, thereby preventing numerical instability caused by out-of-domain inputs.For a deep network consisting of $L$ such layers, the final function approximator $u_{\theta}(x)$ is obtained through layer-by-layer propagation of the input $X^{(0)}=x$:
\begin{equation}
u_{\theta}(x) = (\mathcal{H}^{(L)} \circ \mathcal{H}^{(L-1)} \circ \cdots \circ \mathcal{H}^{(1)})(x).
\end{equation}
This approximator is then embedded into the standard physics-informed framework. By replacing the MLP backbone with KAN, the PIKAN framework significantly enhances the model's expressive capacity, providing a foundation for accurately capturing complex features in PDE solutions.

\subsection{Dynamic Adaptive Loss Weighting Function}
While PINNs have proven effective for many simple problems, they often face significant challenges when applied to more complex ones. A primary contributing factor is that traditional PINNs use fixed weights to balance these loss terms, but their performance is highly dependent on the choice of weights and requires tedious manual tuning. Therefore, selecting appropriate loss weights to balance the various components of the loss function can effectively accelerate the convergence speed and improve the accuracy of PINNs\cite{mcclenny2023selfadaptive}\cite{xiang2022selfadaptive}. Inspired by the uncertainty weighting method in multi-task learning\cite{kendall2018multitask}, we can treat each loss term—which typically includes the PDEs residual, Boundary Conditions, and Initial Conditions—as a separate task. Thus, the process of solving a PDE can be viewed as a multi-task learning problem. By using homoscedastic uncertainty as the basis for loss weighting in a multi-task learning problem, we note that homoscedastic uncertainty is an impromptu uncertainty that does not depend on the input data. It is not a model output but a quantity that remains constant across all input data and varies between different tasks. Therefore, it can be described as task-dependent uncertainty. The implementation process is as follows.
We assume that the PDE solution follows a Gaussian distribution, and its likelihood function can be expressed as:
\begin{equation}
p(u|\hat{u}, \sigma^2) = \frac{1}{\sqrt{2\pi\sigma^2}} \exp\left(-\frac{\|u-\hat{u}\|^2}{2\sigma^2}\right).
\end{equation}
where $\hat{u}$ is the neural network's prediction, and $\sigma^2$ is the variance parameter, representing the degree of task uncertainty.
In PIKANs, we need to simultaneously satisfy the PDE residual, boundary conditions, and initial conditions, which constitutes a multi-task learning problem. Assuming that the tasks are conditionally independent, the joint likelihood function is:
\begin{equation}
p(u|\hat{u}, \sigma_r^2, \sigma_b^2, \sigma_i^2) = \prod_{j\in\{r,b,i\}} \frac{1}{\sqrt{2\pi\sigma_j^2}} \exp\left(-\frac{\mathcal{L}_j}{2\sigma_j^2}\right),
\end{equation}
where $\sigma_r^2, \sigma_b^2, \sigma_i^2$ are the uncertainty parameters corresponding to the PDE residual, boundary conditions, and initial conditions, respectively. $\mathcal{L}_r=\|F(\hat{u})\|^2$ is the PDE residual loss, $\mathcal{L}_b=\|B(\hat{u})\|^2$ is the boundary condition loss, and $\mathcal{L}_i=\|u_0-\hat{u}\|^2$ is the initial condition loss.
By minimizing the negative log-likelihood, we derive the loss function:
\begin{equation}
\begin{split}
& -\log p(u|\hat{u}, \sigma_r^2, \sigma_{b}^2, \sigma_{i}^2) \\
&= -\sum_{j\in\{r,b,i\}}\log \biggl[ \frac{1}{\sqrt{2\pi\sigma_j^2}} \exp \biggl(-\frac{\|u-\hat{u}\|^2}{2\sigma_j^2} \biggr) \biggr] \\
&= \sum_{j\in\{r,b,i\}} \biggl( \frac{1}{2\sigma_j^2} \mathcal{L}_j + \frac{1}{2} \log\sigma_j^2 + \frac{1}{2}\log(2\pi) \biggr).
\end{split}
\end{equation}
Ignoring the constant term $\frac{1}{2}\log(2\pi)$, we obtain the base loss function:
\begin{equation}
\mathcal{L}_{\text{base}}(W,\sigma_r^2, \sigma_b^2, \sigma_i^2) = \sum_{j\in\{r,b,i\}} \left( \frac{1}{2\sigma_j^2} \mathcal{L}_j(W) + \frac{1}{2} \log\sigma_j^2 \right).
\label{eq：loss}
\end{equation}

The first term in Eq.(\ref{eq：loss}) represents the weighted loss function for each task, while the second term is a regularization term that penalizes tasks with high uncertainty. Minimizing $\mathcal{L}_{\text{base}}$ in Eq.(\ref{eq：loss}) is equivalent to maximizing the joint probability distribution of all losses, which facilitates the adaptive allocation of appropriate weight coefficients to the various components of the loss, ensuring a balance among them. The parameter $\sigma_j^2$ represents the uncertainty associated with each task. If a task is easy to learn and exhibits low uncertainty, it will be assigned a higher weight; conversely, a task with high uncertainty will receive a lower weight.
In summary, the uncertainty-based loss function is:
\begin{equation}
\begin{aligned}
\mathcal{L}(u,\sigma) &= \frac{1}{2\sigma_{r}^2} \mathcal{L}_{r}(u)+\frac{1}{2\sigma_{bc}^2} \mathcal{L}_{bc}(u)+\frac{1}{2\sigma_{ic}^2} \mathcal{L}_{ic}(u) \\
&\quad + \frac{1}{2} \log\sigma_r^2+\frac{1}{2}\log\sigma_{bc}^2+\frac{1}{2} \log\sigma_{ic}^2.
\end{aligned}
\end{equation}

However, in multi-task PDE solving, different loss terms (PDE residual, boundary conditions, initial conditions) often have different orders of magnitude and convergence characteristics. In the early stages of training, certain loss terms may dominate the gradient updates. If the gradient magnitude of some loss terms is too large, it can lead to training instability and weight imbalance, causing the weights of some tasks to grow excessively, thereby suppressing the learning of other tasks and causing the model to prematurely fall into local optima for certain tasks. Therefore, inspired by \cite{niu2025improved}, this paper introduces a dynamic upper bound weight constraint method. Its core mathematical expression is:
\begin{equation}
    \gamma(t) = \gamma_{\text{max}} \cdot \exp(-\alpha t) + \gamma_{\text{min}},
\end{equation}
where t is the training epoch number, $\gamma_{\text{max}}$ and $\gamma_{\text{min}}$ are the initial and final values of the upper bound, and $\alpha$ is the decay rate. This constraint achieves a smooth transition from a loose to a strict constraint through a time-dependent decay mechanism.
The adaptive weight calculation based on the dynamic upper bound is:
\begin{equation}
    \lambda_j = \frac{1}{\sigma_j^2 + \frac{1}{\gamma(t)} + \epsilon},
\end{equation}
where $\epsilon=10e^{-12}$ is a small constant added for numerical stability. When $\gamma(t)$ is large, $\frac{1}{\gamma(t)}\rightarrow 0$, and the weight is mainly controlled by $\sigma_j^2$; when $\gamma(t)$ is small, the term $\frac{1}{\gamma(t)}$ becomes significant, constraining the range of weight changes.

Finally, we obtain the extended constrained adaptive loss function:
\begin{equation}
    \mathcal{L}_{\text{adaptive}} = \sum_{j\in\{r,b,i\}} \left( \lambda_j \mathcal{L}_j + \log\left( \sigma_j^2 + \frac{1}{\gamma(t)} \right) \right).
\end{equation}
In the initial stage of training, $\gamma(t) \approx \gamma_{\max}$, the constraint is loose, allowing for larger adjustments in weights, and the model can freely explore the relative importance of each loss term. In the middle stage, $\gamma(t)$ decays exponentially, the constraint gradually tightens, and the optimization process becomes more refined, thereby preventing certain loss terms from excessively dominating. In the later stage, $\gamma(t)\rightarrow\gamma_{\min}$, the weights provide a stable constraint, ensuring the convergence of the equation and preventing divergence in the late training phase.
\subsection{DBAW-PIKAN}
To simultaneously address the limitations in network expressivity and the imbalance in multi-objective optimization within the PINNs framework, we propose an integrated DBAW-PIKAN framework. This framework creatively integrates the high-expressivity PIKAN architecture with the DBAW self-adaptive weighting strategy. 

This integration is not a mere superposition of two independent improvements but is fundamentally rooted in the synergy between them.On one hand, the PIKAN architecture, as constructed in Section 2.2, leverages the superior function approximation capabilities of Kolmogorov-Arnold Networks to provide the potential for accurate PDE solutions, effectively mitigating the spectral bias inherent in traditional MLPs \cite{wang2025kolmogorov}. However, the potential of a high-expressivity architecture cannot be fully realized if the training process is unstable. As an implementation of PINNs, PIKAN still inherits the core challenge of imbalanced training dynamics caused by multi-objective loss functions.On the other hand, the DBAW strategy proposed in Section 2.3 provides stability to the optimization process by dynamically balancing the contributions of various loss components. This effectively prevents certain loss terms from dominating the optimization and suppressing the learning of other physical constraints. However, experimental results (see Section 3.2) indicate that the DBAW strategy is not a universal "plug-and-play" module. When DBAW is combined with a traditional MLP architecture (i.e., the DBAW-PINN model), it exhibits suboptimal performance in solving the nonlinear Burgers equation. This suggests that the efficacy of the DBAW strategy is highly dependent on a compatible and sufficiently expressive underlying network architecture.

Consequently, the DBAW-PIKAN framework is designed to capitalize on this synergy. It exploits the inherent advantages of the PIKAN architecture in function representation and the robustness of the DBAW strategy in multi-objective optimization scenarios. PIKAN provides the necessary foundation for DBAW to apply its stable optimization capabilities, while DBAW ensures the training stability required for PIKAN to fully exert its high expressivity. This combination not only yields significant performance improvements in specific problems but also demonstrates enhanced robustness and higher accuracy when facing complex multi-constraint challenges.

\begin{figure*}[htbp]
    \centering
    \includegraphics[width=0.8\linewidth]{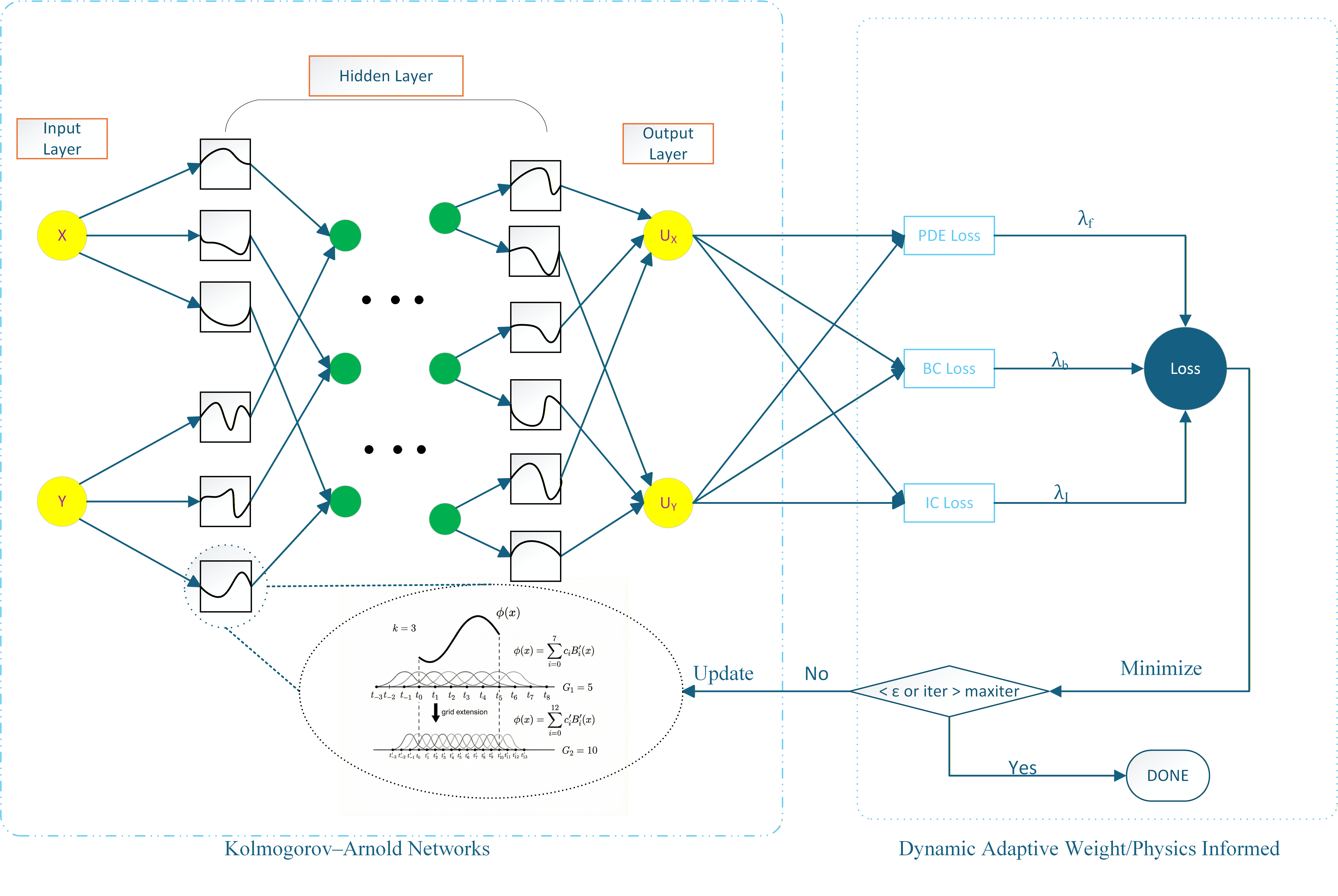}
    \caption{DBAW-PIKAN Framework Diagram}
    \label{fig_DBAW_PIKAN_KUANGJIA}

\end{figure*}

The ultimate DBAW-PIKAN framework is illustrated in Fig. \ref{fig_DBAW_PIKAN_KUANGJIA}. The left side of the architecture comprises the Kolmogorov-Arnold Network, responsible for high-precision function approximation of input coordinates. The right side represents the dynamic self-adaptive physical information component, which calculates physical losses based on the network output and dynamically updates weights via the DBAW strategy to guide the overall optimization. The complete training procedure is detailed in Algorithm 1, illustrating the alternating update process of the model parameters $\theta$ and the uncertainty parameters $\log\sigma$.

\begin{algorithm}[t]
\caption{Dynamic Adaptive Weight PIKAN Algorithm}
\label{alg:DBAW-PIKAN}
\small
\vspace{0.5em}
\begin{tabular}{ll}
\textbf{Input:} & $\gamma_{\max}, \gamma_{\min}, \alpha, N_{int}, N_{bc}, T$ \\
\textbf{Output:} & $\theta^*, \log \sigma_r^*, \log \sigma_u^*$ \\
\end{tabular}
\vspace{0.5em}
\hrule
\vspace{0.8em}

\noindent \textbf{Step1:} Initialize the Kolmogorov-Arnold Network (KAN) model $f_{\theta}$ and adaptive weight parameters $\log \sigma_{r} \leftarrow 0, \log \sigma_{u} \leftarrow 0$. 

\vspace{0.8em}
\noindent \textbf{Step2:} Initialize the optimizers $\text{Adam}_{\theta}$ and $\text{Adam}_{\sigma}$ with designated learning rates for model parameters and weights respectively.

\vspace{0.8em}
\noindent \textbf{Step3:} Execute the training loop for $t=1, \dots, T$: 

\begin{enumerate}[label=(\alph*), leftmargin=2.5em, itemsep=0.8ex, topsep=0.5ex]
    \item Sample interior points $\{(x_i, y_i)\}_{i=1}^{N_{int}}$ and boundary points $\{(x_j^b, y_j^b)\}_{j=1}^{N_{bc}}$. 
    
    \item Compute the dynamic upper bound $\gamma(t) = \gamma_{\max} \cdot \exp(-\alpha t) + \gamma_{\min}$ and evaluate PDE residual loss $\mathcal{L}_{PDE}$ and boundary loss $\mathcal{L}_{BC}$. 
    
    \item Calculate the adaptive weights $\lambda_r$ and $\lambda_u$ as: 
    \begin{equation*}
        \lambda_r = \min \left\{ \frac{1}{\sigma_r^2 + \frac{1}{\gamma(t)}}, \gamma(t) \right\}, \quad \lambda_u = \min \left\{ \frac{1}{\sigma_u^2 + \frac{1}{\gamma(t)}}, \gamma(t) \right\}.
    \end{equation*}

    \item Update the KAN model parameters $\theta$ and weight parameters $\log \sigma$: 
    \begin{align*}
        \theta_{t+1} &= \text{Adam}_{\theta}(\mathcal{L}_{Total}, \theta_{t}) \\
        \log \sigma_{r,t+1}, \log \sigma_{u,t+1} &= \text{Adam}_{\sigma}(\mathcal{L}_{Total}, \log \sigma_{r,t}, \log \sigma_{u,t})
    \end{align*}
\end{enumerate}

\vspace{0.5em}
\noindent \textbf{Return:} The optimized model parameters $\theta^*$ and the final adaptive weights $\log \sigma_{r}^*, \log \sigma_{u}^*$. 
\vspace{0.5em}
\hrule
\end{algorithm}

\section{Experiments}
To quantify the prediction error of the models, we employ the relative $L_2$ error as the performance metric, which is defined as:
\begin{equation}
L_{2}=\frac{\sqrt{\sum_{i=1}^{N}\left|\hat{u}\left(x_{i}, t_{i}\right)-u\left(x_{i}, t_{i}\right)\right|^{2}}}{\sqrt{\sum_{i=1}^{N}\left|u\left(x_{i}, t_{i}\right)\right|^{2}}},
\label{eq:l2_error}
\end{equation}
where $\hat{u}(x_i, t_i)$ represents the predicted value, $u(x_i, t_i)$ is the ground-truth solution, and $N$ denotes the total number of sampling points.

The objective of this section is to evaluate the performance of the proposed DBAW self-adaptive loss strategy and the enhanced PIKAN framework through a series of numerical experiments. The experimental design encompasses the selection of representative partial differential equations, detailed experimental configurations, and the adopted evaluation metrics. A comprehensive comparative analysis will be conducted across four models: the standard PINNs, DBAW-PINN, PIKAN, and the proposed DBAW-PIKAN. Our evaluation primarily focuses on the discrepancies in their training accuracy, convergence rates, and computational efficiency.
\subsection{Klein-Gordon Equation}
The Klein-Gordon equation describes the dynamics of spinless particles within the frameworks of relativistic quantum mechanics and quantum field theory.  As a nonlinear partial differential equation, it finds extensive applications in fields such as particle physics and nonlinear optics. The Klein-Gordon equation is formulated as:
\[
\begin{cases}
u_{tt} + \alpha \Delta u + \beta u + \gamma u^{k} = f(x,t), & (x,t) \in \Omega \times [0,T], \\ 
u(x,0) = g_{1}(x), & x \in \Omega, \\ 
u_{t}(x,0) = g_{2}(x), & x \in \Omega, \\ 
u(x,t) = h(x,t), & (x,t) \in \partial \Omega \times [0,T],
\end{cases}
\]
where $\alpha, \beta, \gamma$, and $k$ are known constants; $\Delta$ denotes the Laplacian operator acting on the spatial variables, and $k$ represents the degree of nonlinearity. The functions $f(x,t), g_1(x), g_2(x)$, and $h(x,t)$ are specified conditions, while $u(x,t)$ is the unknown function to be determined.

In this study, the computational domain is set to $\Omega = [0,1]$ and the time interval is $T = [0, 1]$. The equation parameters are chosen as $\alpha = -1, \beta = 0, \gamma = 1$, and $k = 3$. To evaluate the accuracy of the models, we employ a manufactured solution:
\begin{equation}
u(x,t) = x \cos(5\pi t) + (xt)^{3}.
\end{equation}
The corresponding forcing term $f(x,t)$, along with the initial conditions $g_1(x), g_2(x)$ and boundary conditions $h(x,t)$, are analytically derived from this manufactured solution.

\begin{figure*}[htbp]
    \centering
    \begin{subfigure}{\textwidth}
        \centering
        \begin{minipage}{0.32\textwidth}
            \centering
            \includegraphics[width=0.95\linewidth, height=4cm]{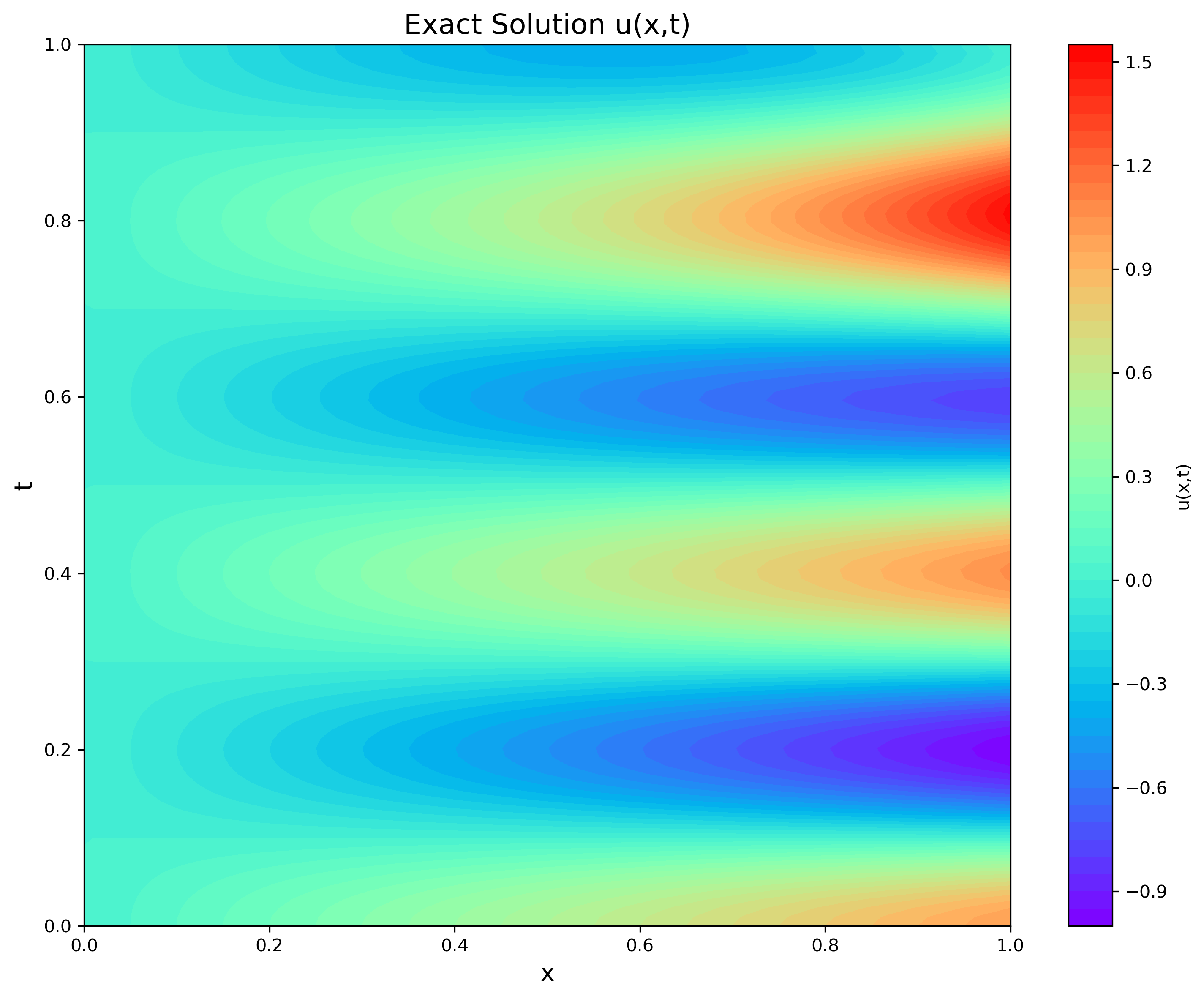}
            \label{fig:pinn_exact}
        \end{minipage}\hfill
        \begin{minipage}{0.32\textwidth}
            \centering
            \includegraphics[width=0.95\linewidth, height=4cm]{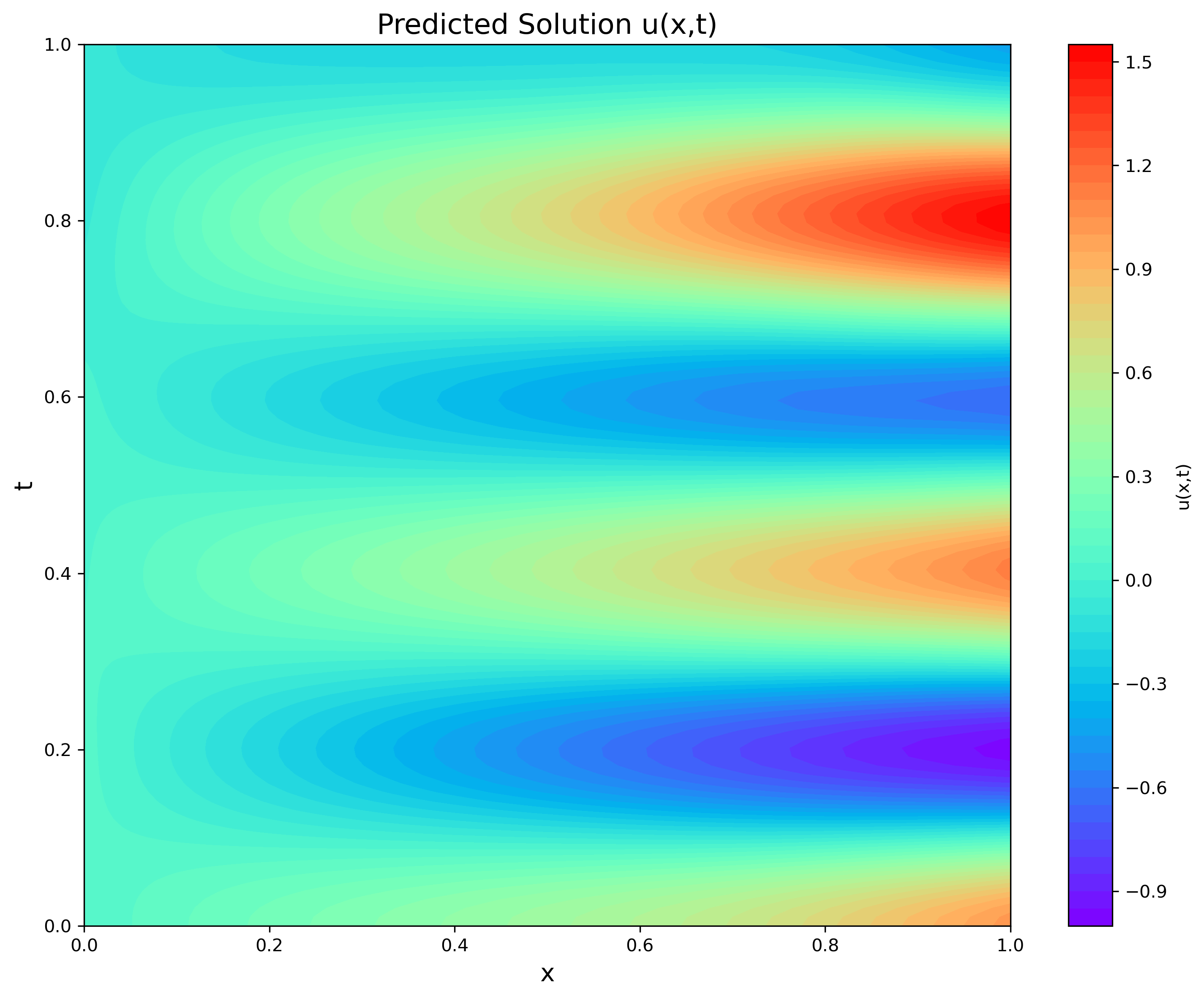}
            \label{fig:pinn_predicted}
        \end{minipage}\hfill
        \begin{minipage}{0.32\textwidth}
            \centering
            \includegraphics[width=0.95\linewidth, height=4cm]{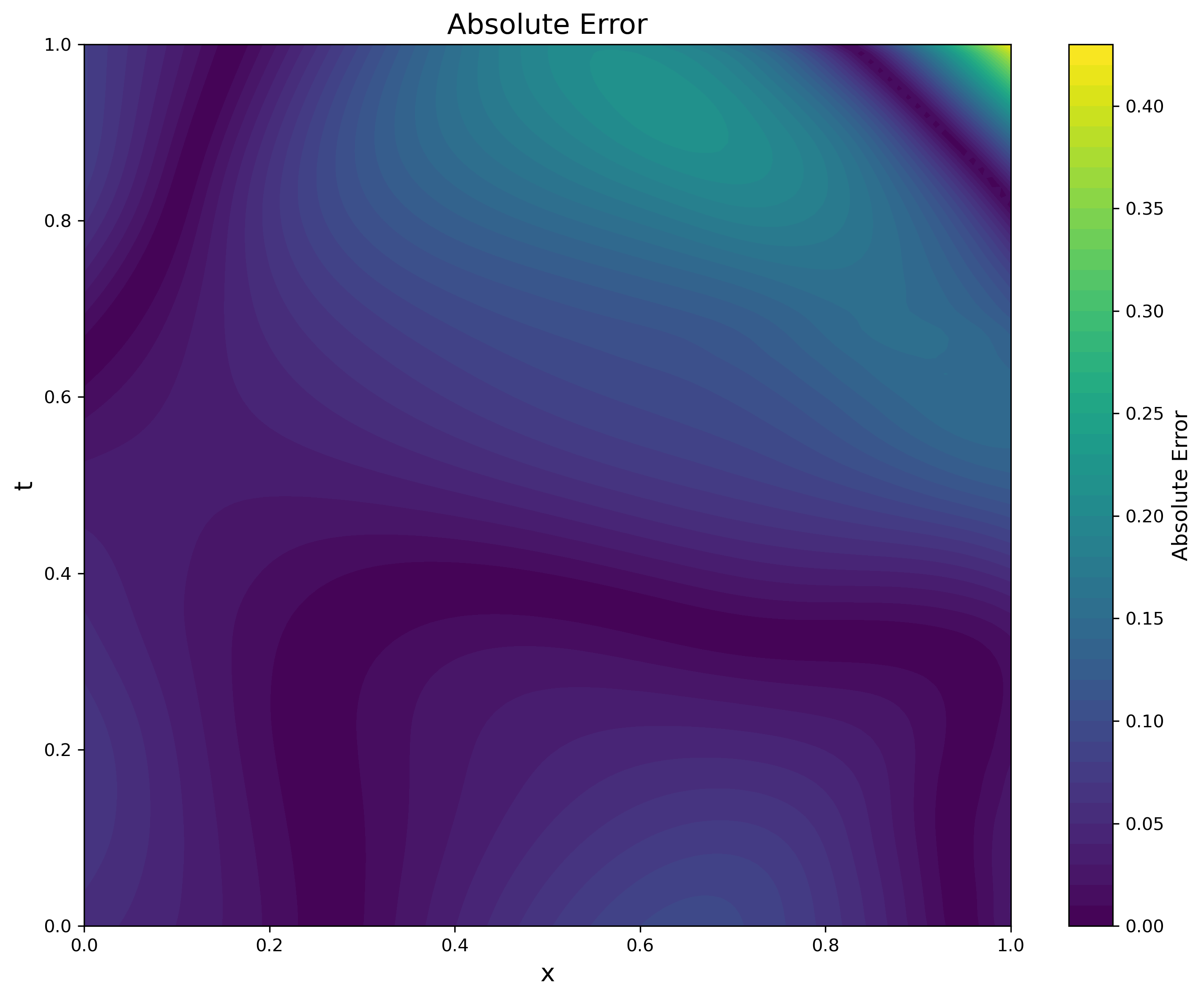}
            \label{fig:pinn_error}
        \end{minipage}
        \caption*{(a) PINN}
    \end{subfigure}
    
    \vspace{0.3cm}
    
    \begin{subfigure}{\textwidth}
        \centering
        \begin{minipage}{0.32\textwidth}
            \centering
            \includegraphics[width=0.95\linewidth, height=4cm]{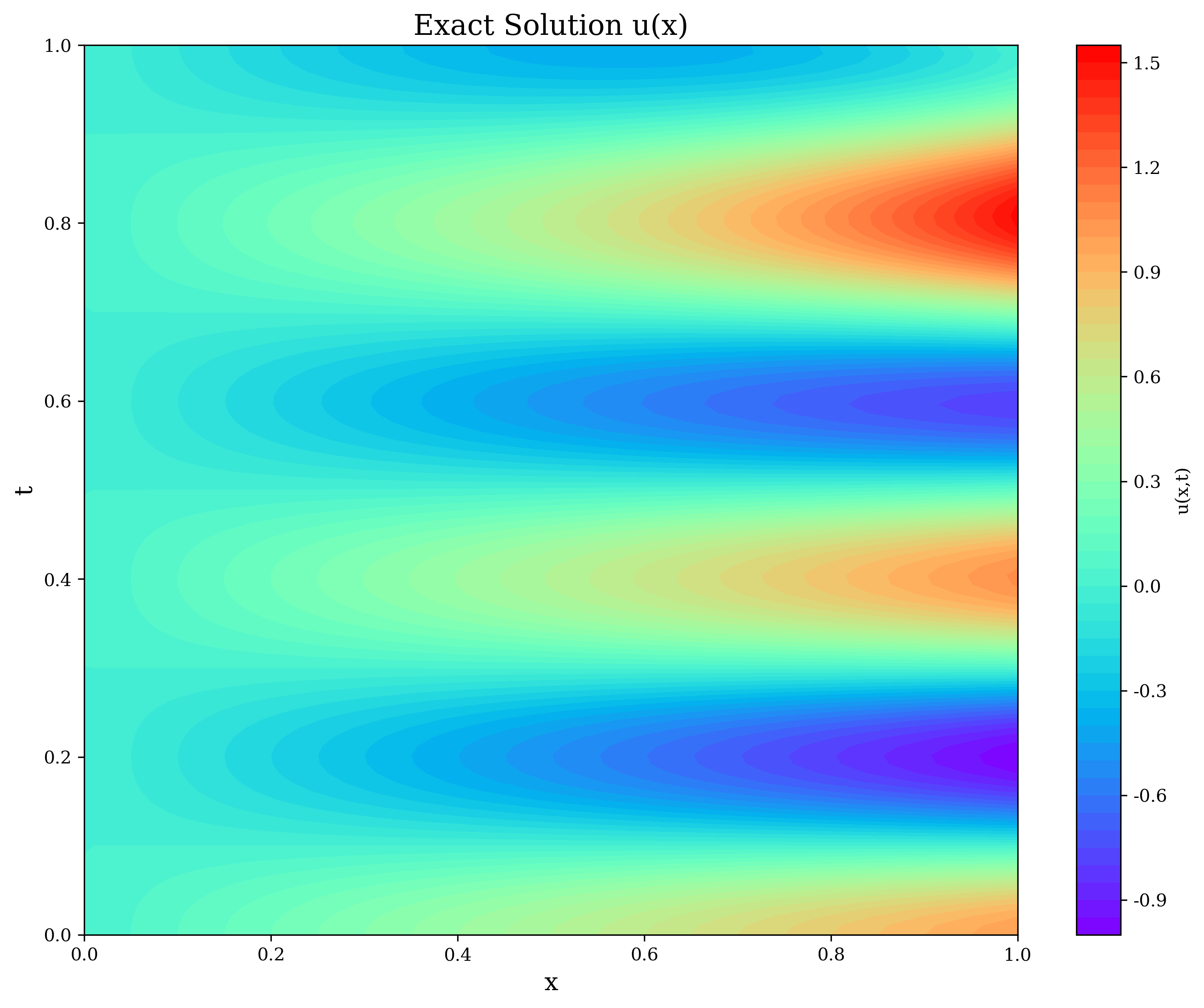}
            \label{fig:dbaw_pinn_exact}
        \end{minipage}\hfill
        \begin{minipage}{0.32\textwidth}
            \centering
            \includegraphics[width=0.95\linewidth, height=4cm]{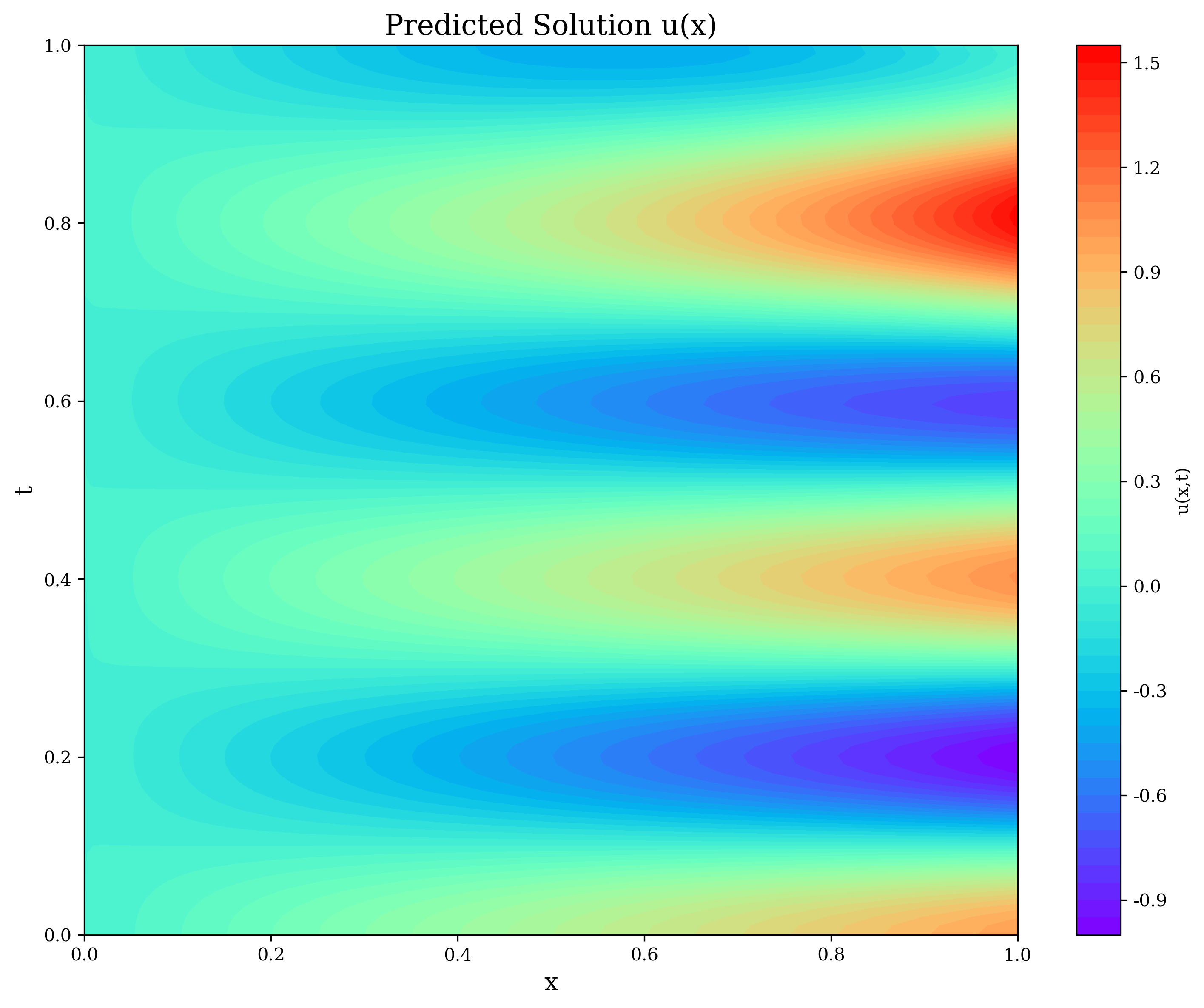}
            \label{fig:dbaw_pinn_predicted}
        \end{minipage}\hfill
        \begin{minipage}{0.32\textwidth}
            \centering
            \includegraphics[width=0.95\linewidth, height=4cm]{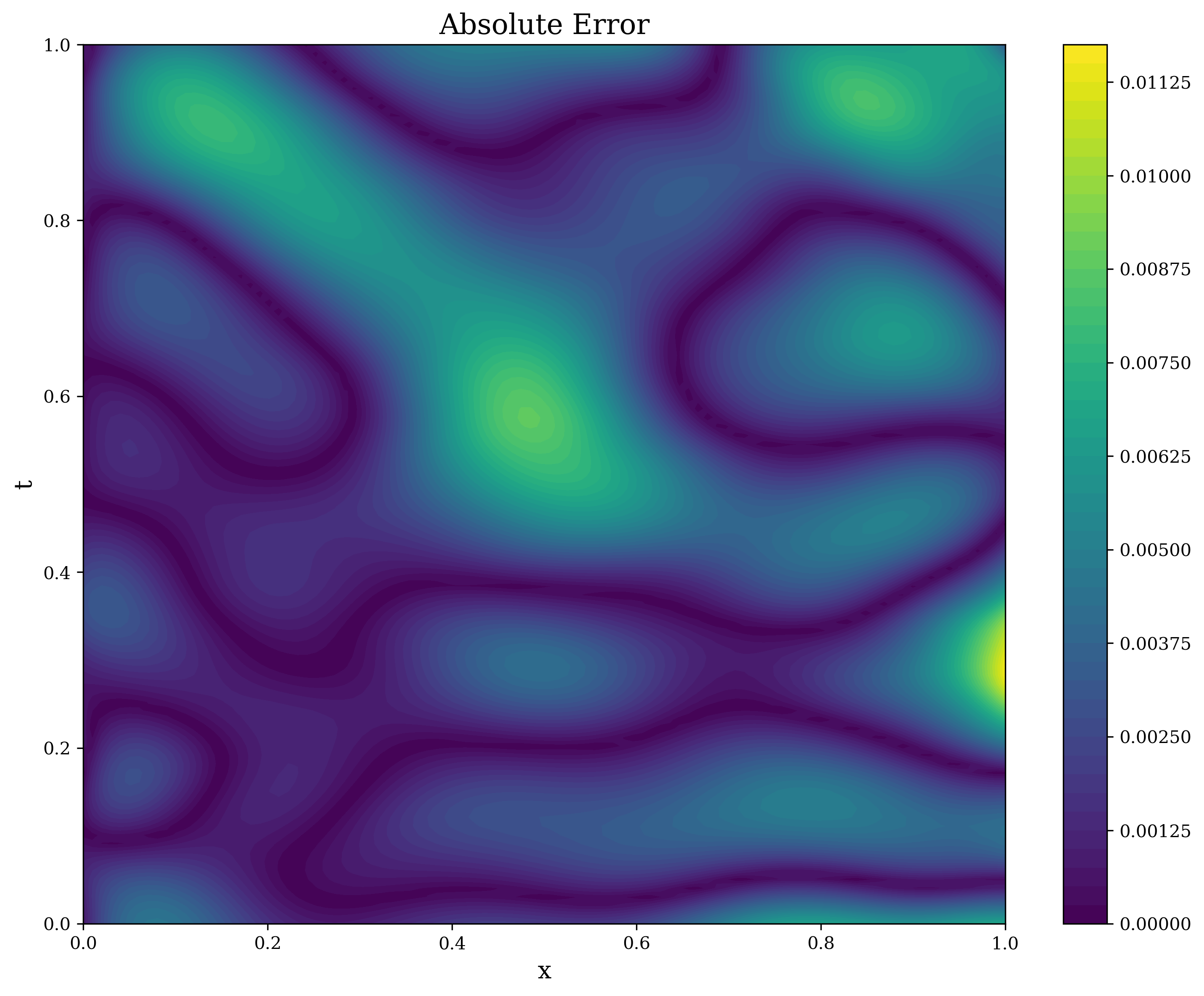}
            \label{fig:dbaw_pinn_error}
        \end{minipage}
        \caption*{(b) DBAW-PINN}
    \end{subfigure}
    
    \vspace{0.3cm}
    
    \begin{subfigure}{\textwidth}
        \centering
        \begin{minipage}{0.32\textwidth}
            \centering
            \includegraphics[width=0.95\linewidth, height=4cm]{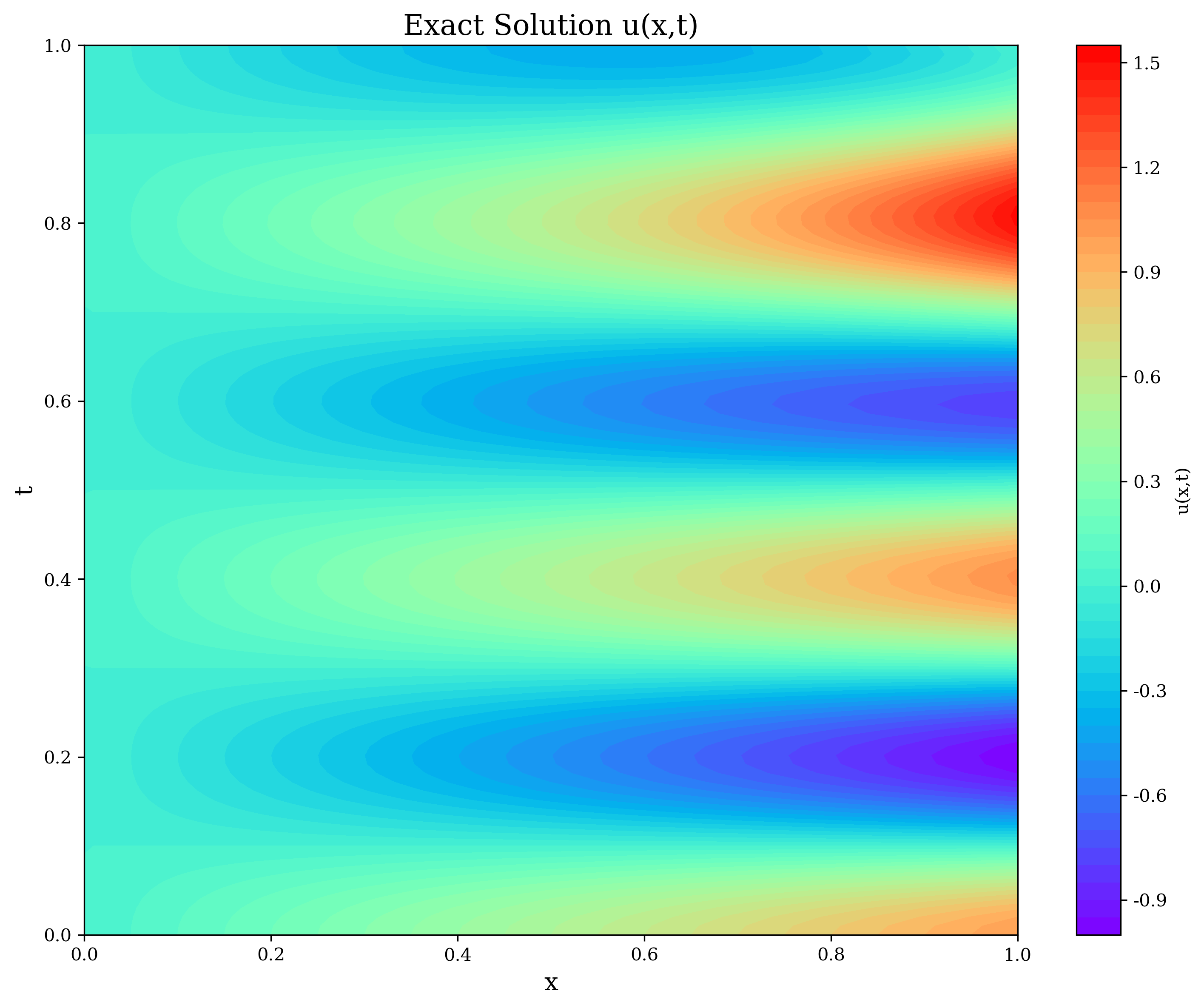}
            \label{fig:pikan_exact}
        \end{minipage}\hfill
        \begin{minipage}{0.32\textwidth}
            \centering
            \includegraphics[width=0.95\linewidth, height=4cm]{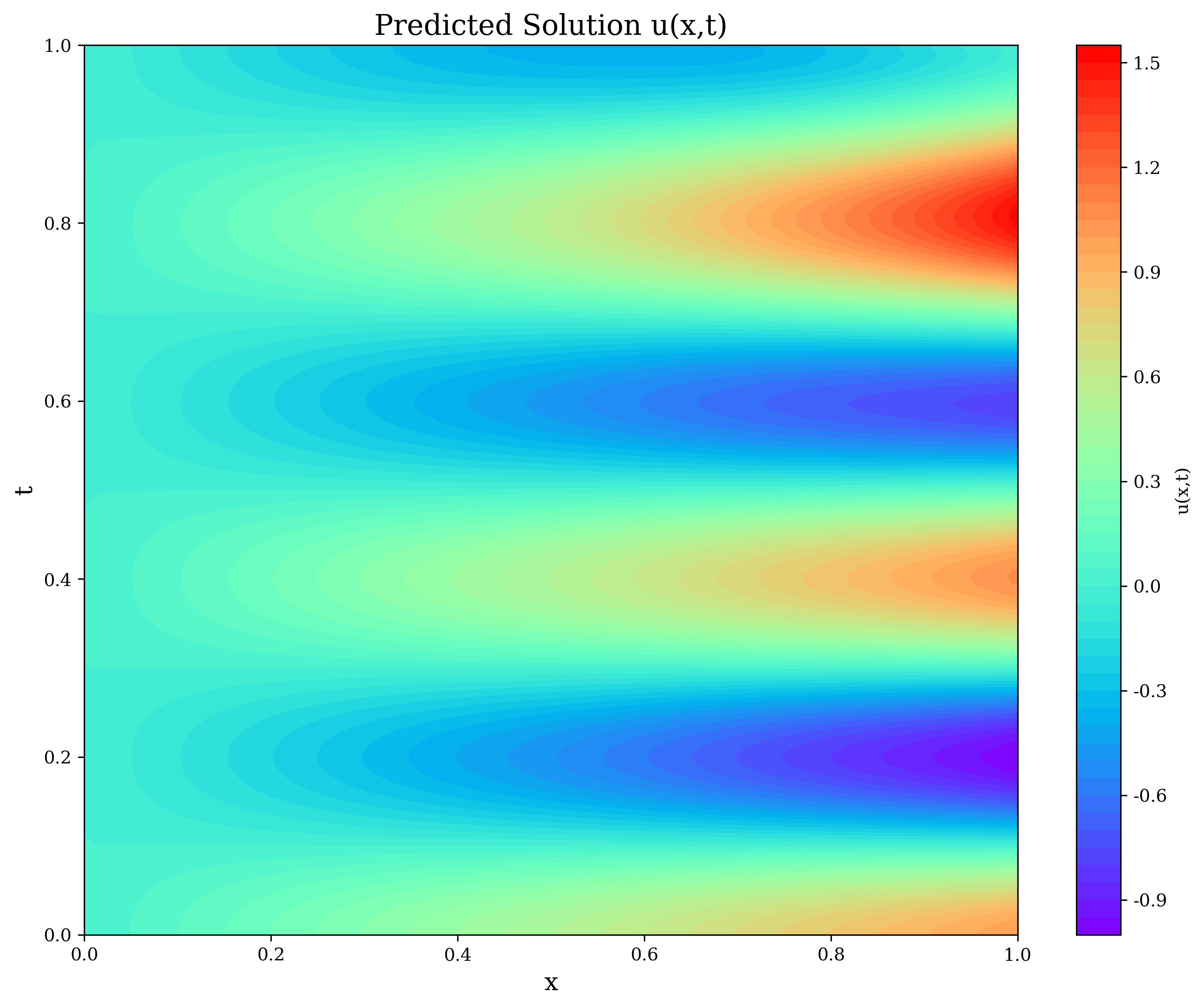}
            \label{fig:pikan_predicted}
        \end{minipage}\hfill
        \begin{minipage}{0.32\textwidth}
            \centering
            \includegraphics[width=0.95\linewidth, height=4cm]{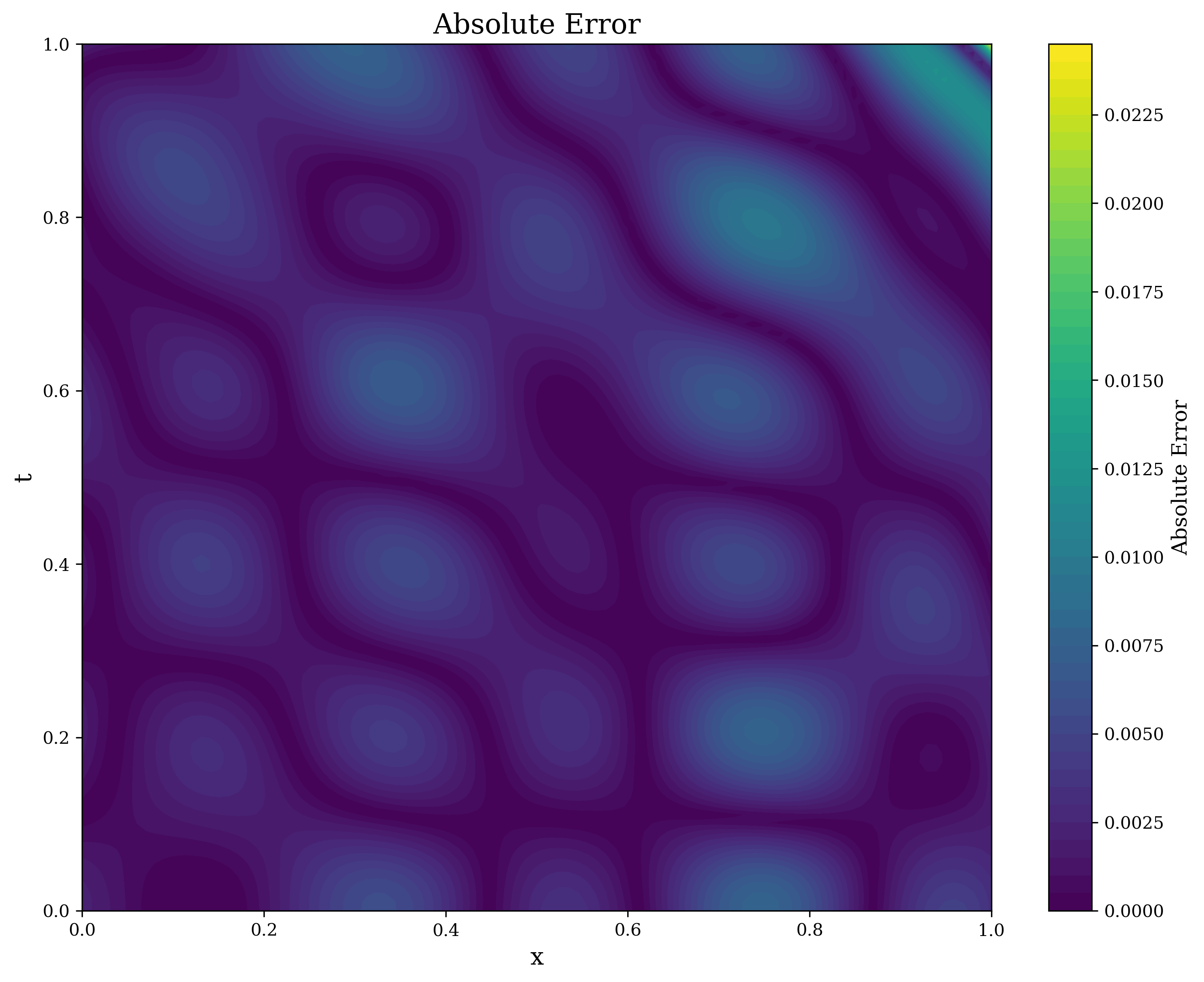}
            \label{fig:pikan_error}
        \end{minipage}
        \caption*{(c) PIKAN}
    \end{subfigure}
    
    \vspace{0.3cm}
    
    \begin{subfigure}{\textwidth}
        \centering
        \begin{minipage}{0.32\textwidth}
            \centering
            \includegraphics[width=0.95\linewidth, height=4cm]{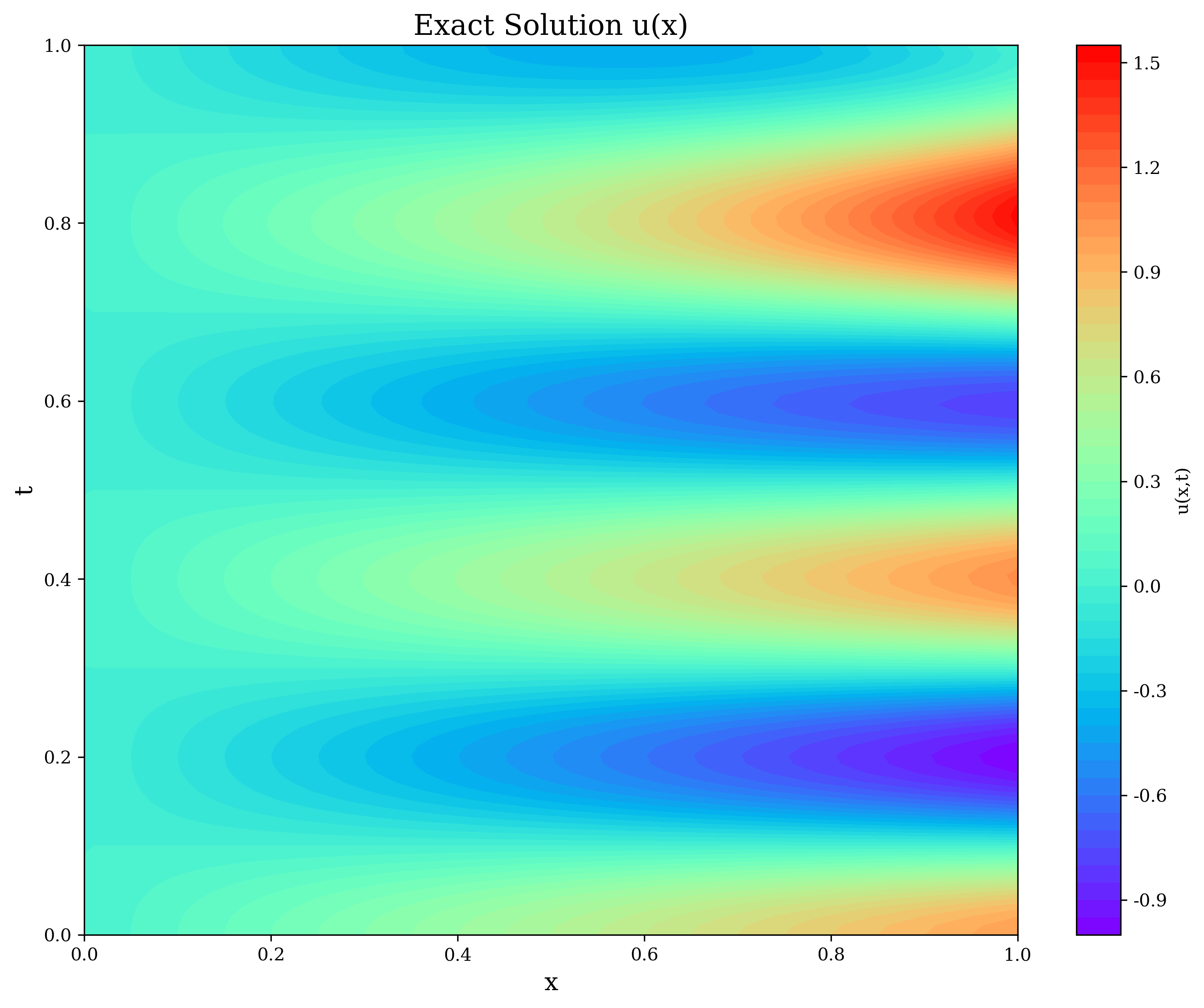}
            \label{fig:dbaw_pikan_exact}
        \end{minipage}\hfill
        \begin{minipage}{0.32\textwidth}
            \centering
            \includegraphics[width=0.95\linewidth, height=4cm]{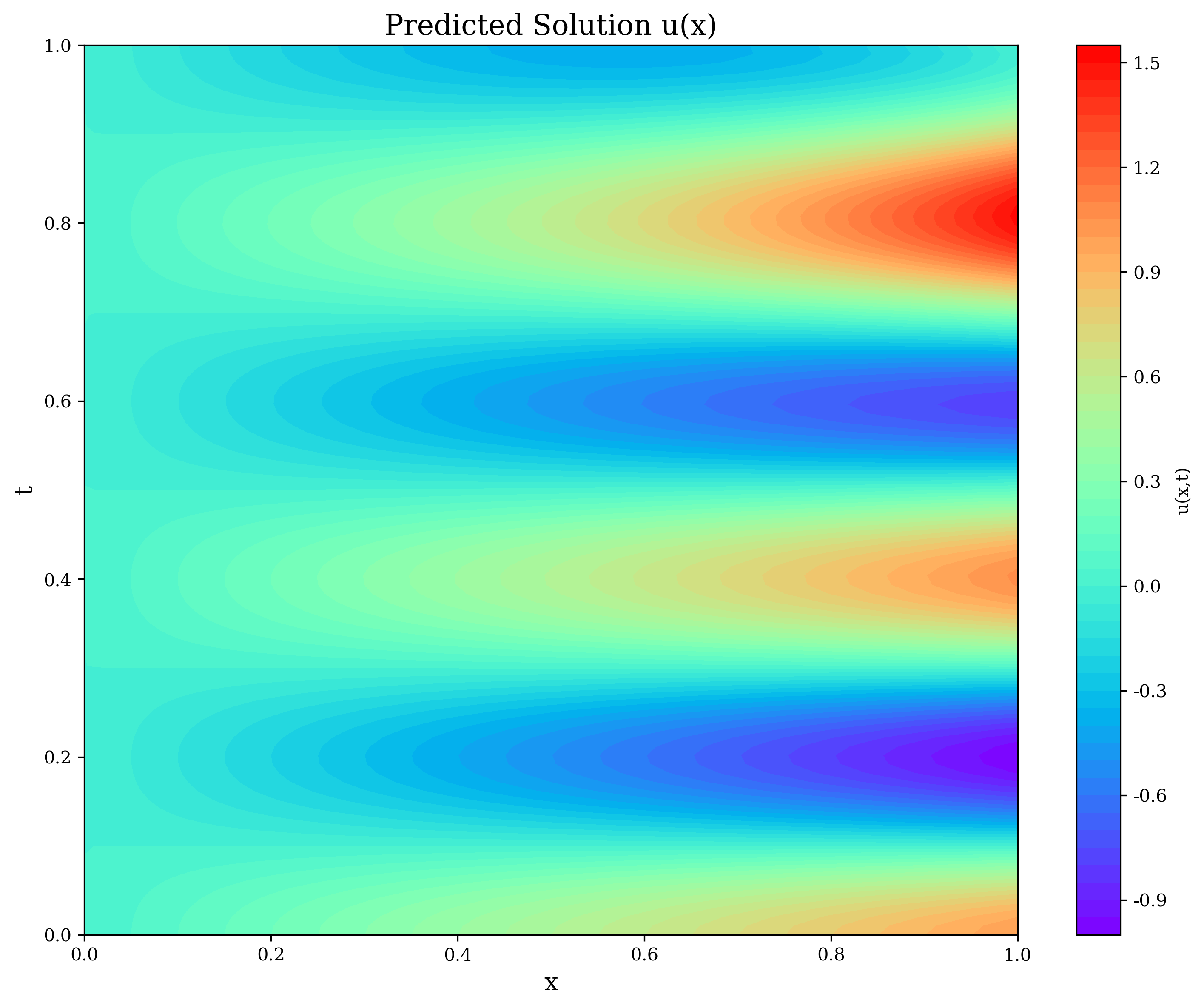} 
            \label{fig:dbaw_pikan_predicted}
        \end{minipage}\hfill
        \begin{minipage}{0.32\textwidth}
            \centering
            \includegraphics[width=0.95\linewidth, height=4cm]{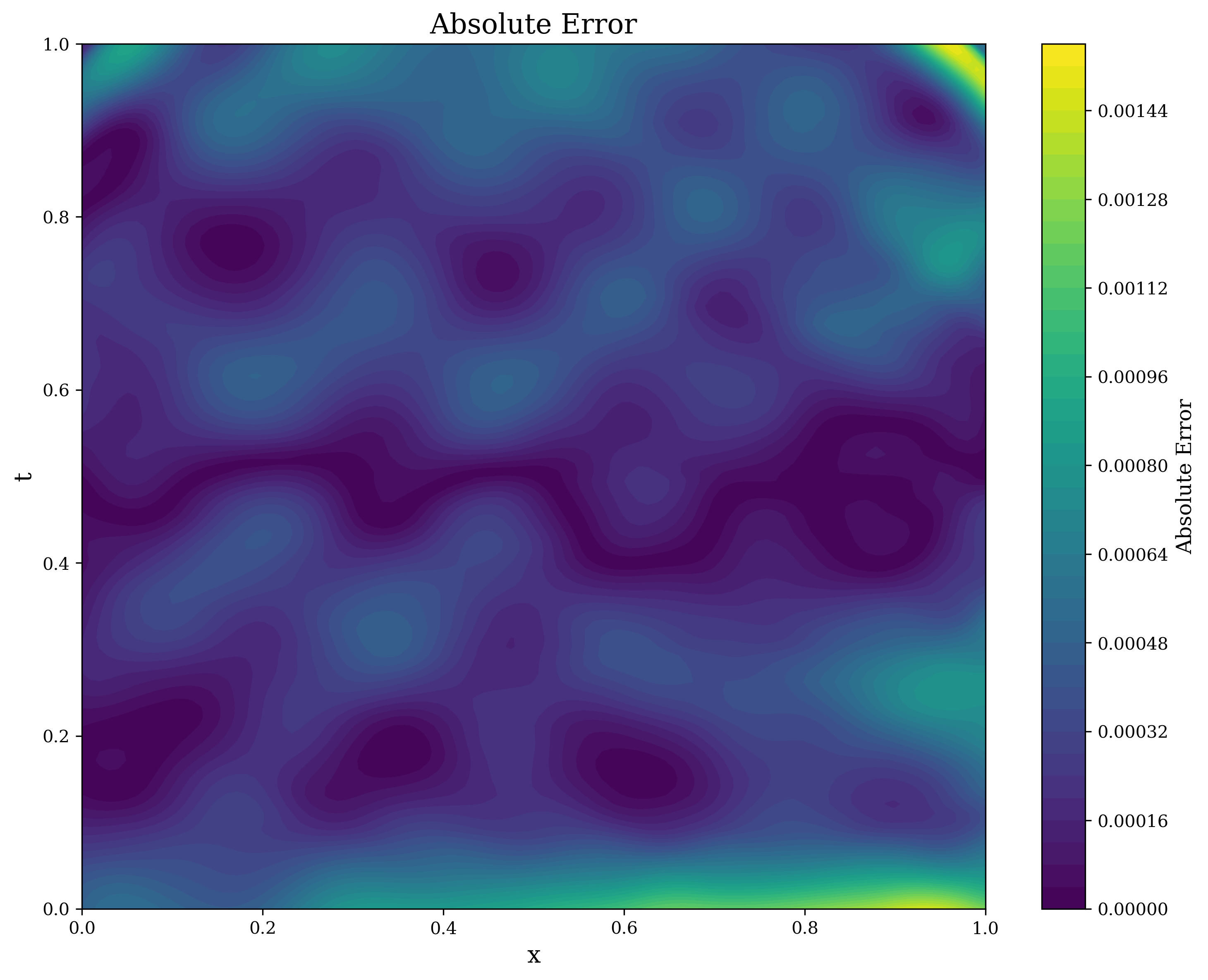}
            \label{fig:dbaw_pikan_error}
        \end{minipage}
        \caption*{(d) DBAW-PIKAN}
    \end{subfigure}
    
    \caption{Numerical solution comparisons of the Klein-Gordon equation among four different methods. Each row displays (from left to right) the exact solution, the predicted solution, and the point-wise absolute error distribution for each respective framework.}
    \label{fig:kg_comparison}
\end{figure*}

We compare the performance of four models: PINN, DBAW-PINN, PIKAN, and DBAW-PIKAN. For the MLP-based architectures (PINN and DBAW-PINN), the network consists of 7 hidden layers with 64 neurons each. For the KAN-based architectures (PIKAN and DBAW-PIKAN), the structure is configured with 4 hidden layers, each with a width of 20, a B-spline order of 4, and a grid size of 20. All models are trained using the Adam optimizer for 50,000 iterations.
Table \ref{tab:klein_gordon_comparison} provides a detailed performance comparison of the four methods for solving the Klein-Gordon equation. Quantitatively, the DBAW-PIKAN method achieves the lowest relative $L_2$ error of $8.88 \times 10^{-4}$. In contrast, PINN yields the highest error at $3.05 \times 10^{-2}$.
The introduction of the Dynamic Balancing Adaptive Weighting (DBAW) strategy reduces the error of DBAW-PINN to $1.17 \times 10^{-2}$, validating the effectiveness of the self-adaptive weighting mechanism. Furthermore, when the architecture is replaced by KAN, the PIKAN model further reduces the error to $7.73 \times 10^{-3}$, showcasing the superior function approximation capabilities of KAN. Ultimately, the DBAW-PIKAN model, which integrates both improvements, achieves a precision nearly an order of magnitude higher than that of PIKAN. This fully demonstrates the synergistic effect between the PIKAN architecture and the DBAW strategy.
\begin{table*}[htbp]
\centering
\caption{Performance comparison of different methods for solving the Klein-Gordon equation.}
\label{tab:klein_gordon_comparison}
\small
\begin{tabular}{lccccccc}
\toprule
\textbf{Method} & \textbf{Order} & \textbf{Grid Size} & \textbf{Architecture} & \textbf{Parameters} & \textbf{Optimizer} & \textbf{Iterations} & \textbf{Relative $L_2$ Error} \\
\midrule
PINN & -- & -- & [2,64,64,64,64,64,64,1] & 21,057 & Adam & 50,000 & $3.05 \times 10^{-2}$ \\
DBAW-PINN & -- & -- & [2,64,64,64,64,64,64,1] & 21,057 & Adam & 50,000 & $8.04 \times 10^{-3}$ \\
PIKAN & 4 & 20 & [2,20,20,20,1] & 22,360 & Adam & 50,000 & $7.61 \times 10^{-3}$ \\
\textbf{DBAW-PIKAN} & 4 & 20 & [2,20,20,20,1] & 22,360 & Adam & 50,000 & $\mathbf{8.88 \times 10^{-4}}$ \\ 
\bottomrule
\end{tabular}

\vspace{0.2cm}
\footnotesize\textit{Note: PINN and DBAW-PINN do not utilize spline basis functions; thus, the order and grid size parameters are not applicable. The best results are highlighted in bold.}
\end{table*}
Fig. \ref{fig:kg_comparison} provides a visual comparison of the results obtained by the four methods. Each row corresponds to a specific method. From left to right, the columns display the exact solution, the predicted solution by the model, and the distribution of the point-wise absolute error between them.

The qualitative comparison results are as follows: (a) The predicted solution of the standard PINN exhibits noticeable visual discrepancies compared to the exact solution, with the absolute error plot showing a widespread error distribution. (b) The prediction results of DBAW-PINN show marked improvement, with a significant reduction in error magnitude, although localized regions of prominent error still persist. (c) The predicted solution of PIKAN is generally closer to the ground-truth solution, and the overall color tone of its error plot (corresponding to lower error values) indicates a widespread reduction in error. (d) The predicted solution of the DBAW-PIKAN method is in high visual agreement with the exact solution. Its absolute error plot features the deepest color tones (representing minimal error values) and a more uniform, subtle error distribution. This observation is highly consistent with the superior numerical accuracy reported in Table \ref{tab:klein_gordon_comparison}.
To further investigate the underlying mechanisms behind the significant performance enhancement achieved by the DBAW-PIKAN framework, this section provides a visual analysis of the adaptive weights and the dynamic evolution of various loss components during the solving of the Klein-Gordon equation.

\begin{figure}[htbp]
  \centering
  \begin{subfigure}[b]{\linewidth}
    \centering
    \includegraphics[width=\linewidth]{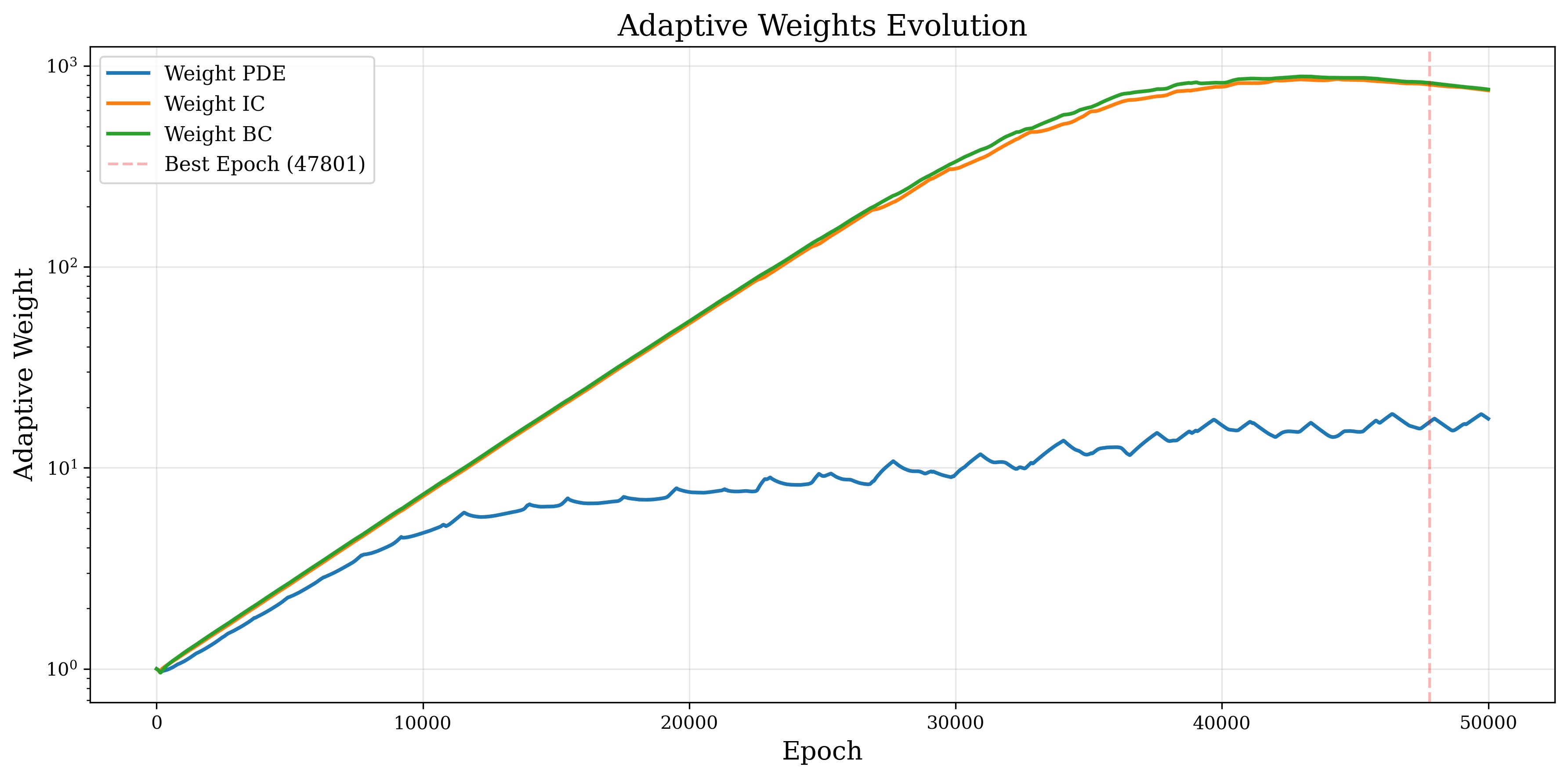}
    \caption{Evolution of adaptive weights}
    \label{fig:kg_weights_evolution} 
  \end{subfigure}
  
  \vspace{1.5em} 

  \begin{subfigure}[b]{\linewidth}
    \centering
    \includegraphics[width=\linewidth]{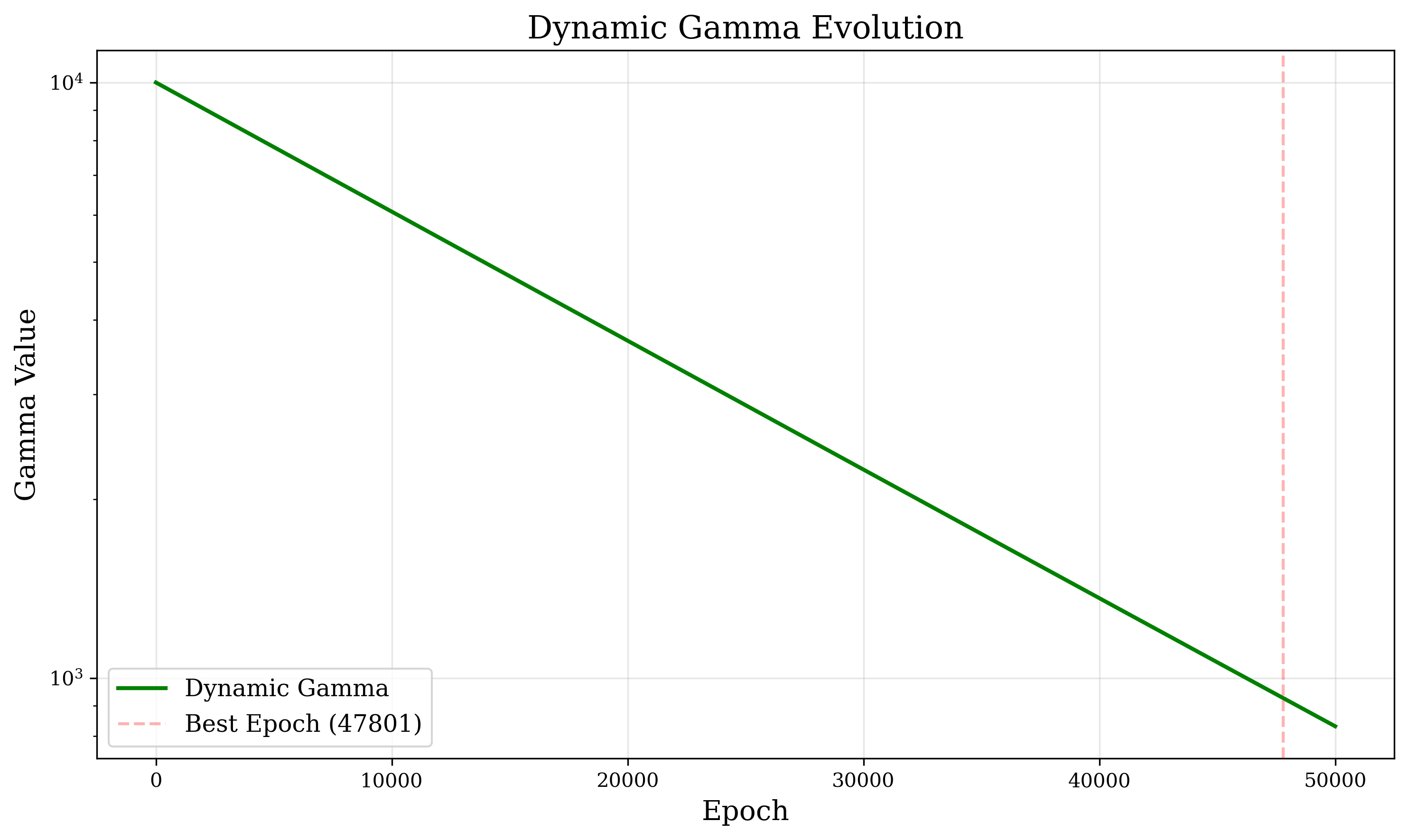}
    \caption{Evolution of the dynamic upper bound $\gamma(t)$}
    \label{fig:kg_gamma_evolution}
  \end{subfigure}

  \vspace{1.5em}

  \begin{subfigure}[b]{\linewidth}
    \centering
    \includegraphics[width=\linewidth]{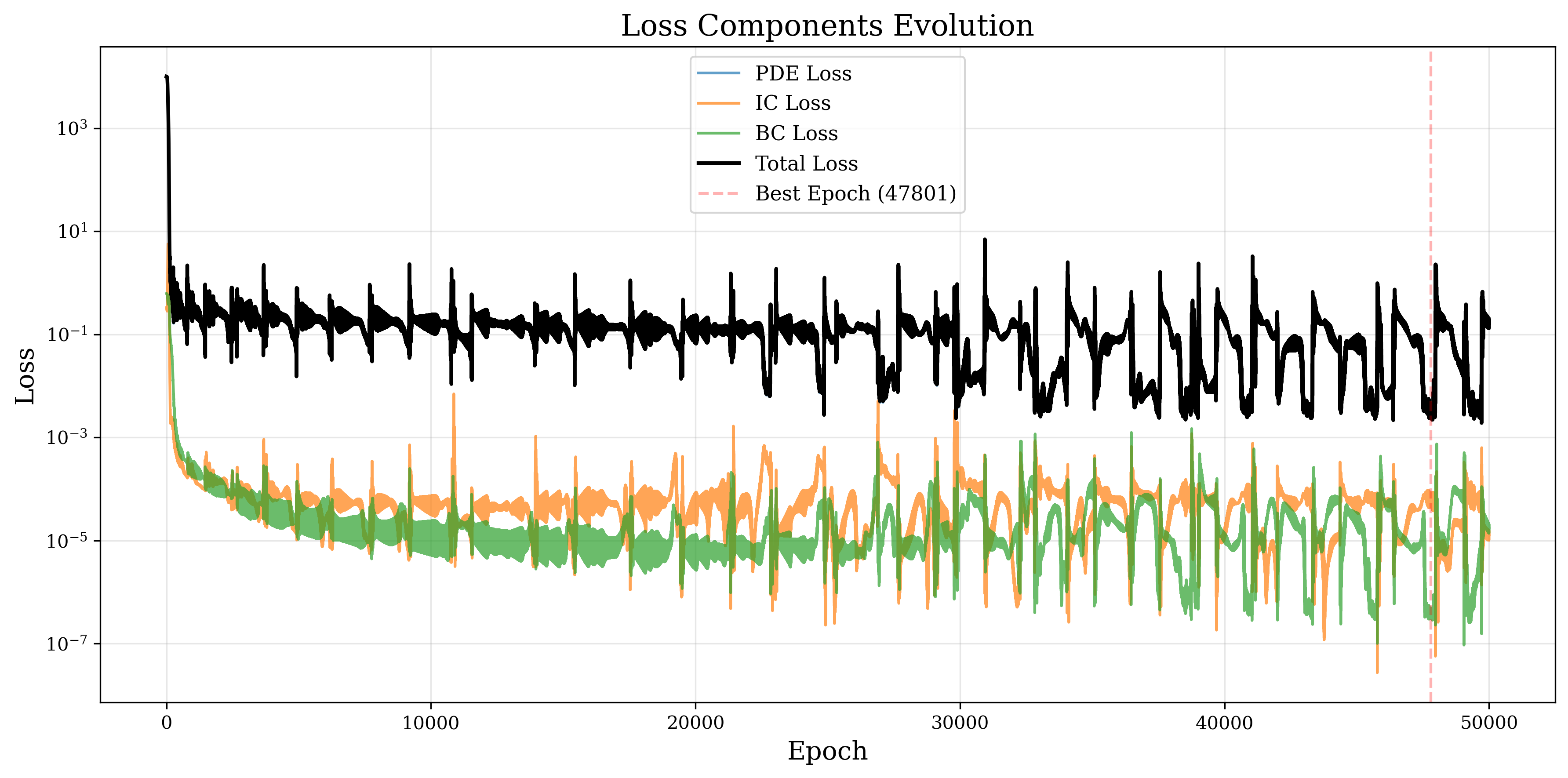}
    \caption{Evolution of loss components}
    \label{fig:loss_components_evolution}
  \end{subfigure}

  \caption{Parameter evolution during the training process for the Klein-Gordon equation: (a) Evolution of adaptive weights ($\lambda_r$, $\lambda_{ic}$, $\lambda_{bc}$) over training epochs; (b) Evolution of the dynamic upper bound parameter $\gamma(t)$ over training epochs; (c) Evolution of individual loss terms over training epochs. The y-axis is on a logarithmic scale, and the vertical dashed line indicates the epoch at which the best performance was achieved.}
  \label{fig:kg_evolution}
\end{figure}

Fig. \ref{fig:kg_weights_evolution} and Fig. \ref{fig:kg_gamma_evolution} illustrate the evolution trajectories of the weights for each loss term ($\lambda_r, \lambda_{ic}, \lambda_{bc}$) and the dynamic constraint parameter $\gamma(t)$ across training epochs, respectively. Complementarily, Fig. \ref{fig:loss_components_evolution} displays the actual evolution of the PDE residual loss ($\mathcal{L}_r$), initial condition loss ($\mathcal{L}_{ic}$), boundary condition loss ($\mathcal{L}_{bc}$), and total loss ($\mathcal{L}_{Total}$) under the influence of these dynamic weights. Together, these figures reveal how the DBAW-PIKAN method achieves a balanced and efficient optimization process through its dynamic weight regulation mechanism.

As previously analyzed based on Fig. \ref{fig:kg_weights_evolution}, the weights $\lambda_{ic}$ and $\lambda_{bc}$ increase rapidly during the early stages of training, reflecting the relative ease with which the network fits the initial and boundary conditions. Without proper constraints, this rapid escalation could potentially suppress the learning of the PDE residual terms.
Fig. \ref{fig:loss_components_evolution} corroborates this observation from the perspective of loss values: $\mathcal{L}_{ic}$ and $\mathcal{L}_{bc}$ (indicated by the orange and green curves) drop rapidly by several orders of magnitude shortly after the start of training. Their convergence rates are significantly faster, and their absolute magnitudes are markedly lower than those of $\mathcal{L}_r$ (blue curve). This highlights the crucial role of introducing the dynamic upper bound $\gamma(t)$. As shown in Fig. \ref{fig:kg_gamma_evolution}, $\gamma(t)$ decays exponentially throughout the training process, imposing a progressively tightening ceiling on the weights. Consequently, as illustrated in Fig. \ref{fig:kg_weights_evolution}, the growth of $\lambda_{ic}$ and $\lambda_{bc}$ is effectively restricted to the order of $10^3$. The direct impact of this constraint is evident in Fig. \ref{fig:loss_components_evolution}: although $\mathcal{L}_{ic}$ and $\mathcal{L}_{bc}$ have converged to low levels (fluctuating between $10^{-5}$ and $10^{-7}$), their weights do not exhibit unbounded growth. This ensures that $\mathcal{L}_r$ (which fluctuates between $10^{-3}$ and $10^{-5}$) maintains a non-negligible proportion of the total loss $\mathcal{L}_{Total}$ (black curve), even with its larger magnitude and slower convergence. As a result, its corresponding weight $\lambda_r$ continues to grow in the later stages (see Fig. \ref{fig:kg_weights_evolution}), allowing the PDE residual term to be persistently minimized by the optimizer.

Accordingly, the superiority of the DBAW-PIKAN method is clearly demonstrated through the synergistic evolution of weights and losses:
\begin{itemize}
\item Achievement of balanced optimization: Compared to standard adaptive methods that may lead to weight imbalance, the dynamic upper bound $\gamma(t)$ ensures that all physical constraints (PDE, IC, and BC) receive sufficient and continuous optimization, despite vast differences in the magnitudes and convergence rates of the loss components. Fig. \ref{fig:loss_components_evolution} shows that all loss components eventually stabilize at low levels, with the total loss decreasing and reaching a steady state.
\item Facilitation of stable convergence and high precision: By preventing weight imbalance and potential optimization oscillations, the DBAW strategy guides the model toward a stable convergence that simultaneously satisfies all physical constraints. The "Best Epoch" (47,801) marked in Fig. \ref{fig:loss_components_evolution} aligns with the markers in the weight evolution plots, indicating that the model achieves its minimum generalization error ($L_2$ error of $8.88 \times 10^{-4}$ in Table \ref{tab:klein_gordon_comparison}) when the weights are relatively stable and the loss components are well-balanced.
\item Synergy with the KAN architecture: The powerful representation capability of the KAN network, combined with the stable optimization path provided by the DBAW strategy, enables DBAW-PIKAN to accurately capture the complex nonlinear solutions of the Klein-Gordon equation.
\end{itemize}

In summary, Figs. \ref{fig:kg_weights_evolution}, \ref{fig:kg_gamma_evolution}, and \ref{fig:loss_components_evolution} collectively provide a robust mechanistic explanation for the superior performance of the DBAW-PIKAN method in solving the Klein-Gordon equation from the perspectives of optimization dynamics and loss composition. By introducing the dynamically decaying upper bound $\gamma(t)$, we successfully achieve a more stable and balanced multi-objective optimization trajectory, which is the key factor behind the significant accuracy improvements over baseline models.

\subsection{Burgers}
The Burgers equation is a fundamental nonlinear partial differential equation (PDE) widely used to describe the dynamic characteristics of viscous fluids. It has extensive applications in nonlinear acoustics, gas dynamics, and fluid mechanics. The governing equation is formulated as:
\begin{equation}
\frac{\partial u}{\partial t} + u \frac{\partial u}{\partial x} = \nu \frac{\partial^{2} u}{\partial x^{2}}
\end{equation}
subject to the following initial and boundary conditions:
\begin{align*}
u(x,0) &= g(x), \quad x \in \Omega, \\
u(-1,t) &= h_1(t), \quad t \in [0,T], \\
u(1,t) &= h_2(t), \quad t \in [0,T].
\end{align*}
While analytical solutions for this equation are generally unavailable, numerical approximations can be obtained using various computational methods \cite{raissi2019physics}.

Consistent with the previous experimental configurations, we evaluate four models: PINN, DBAW-PINN, PIKAN, and DBAW-PIKAN. Specifically, the MLP-based models (PINN and DBAW-PINN) utilize an architecture of [2, 64, 64, 64, 64, 64, 64, 1], while the KAN-based models (PIKAN and DBAW-PIKAN) employ a [2, 20, 20, 20, 1] structure with a B-spline order of 4 and a grid size of 20. All models are trained for 50,000 iterations using the Adam optimizer.
Table \ref{tab:burgers_comparison} summarizes the quantitative results for the Burgers equation. The standard PINN achieves a relative $L_2$ error of $1.23 \times 10^{-3}$. Notably, DBAW-PINN exhibits suboptimal performance on this problem, with the error escalating to $2.71 \times 10^{-1}$. This suggests that for certain nonlinear problems, adaptive weighting strategies may interfere with the training process if not properly compatible with the specific MLP architecture.

\begin{table*}[!t]
\centering
\caption{Performance comparison of different methods for solving the Burgers equation. }
\label{tab:burgers_comparison}
\small
\begin{tabular}{lccccccc}
\toprule
\textbf{Method} & \textbf{Order} & \textbf{Grid Size} & \textbf{Architecture} & \textbf{Parameters} & \textbf{Optimizer} & \textbf{Iterations} & \textbf{Relative $L_2$ Error} \\
\midrule
PINN & -- & -- & [2,64,64,64,64,64,64,1] & 21,057 & Adam & 50,000 & $1.23 \times 10^{-3}$  \\
DBAW-PINN & -- & -- & [2,64,64,64,64,64,64,1] & 21,057 & Adam & 50,000 & $2.71 \times 10^{-1}$  \\
PIKAN & 4 & 20 & [2,20,20,20,1] & 22,360 & Adam & 50,000 & $1.16 \times 10^{-3}$  \\
\textbf{DBAW-PIKAN} & 4 & 20 & [2,20,20,20,1] & 22,360 & Adam & 50,000 & $\mathbf{4.70 \times 10^{-4}}$ \\ 
\bottomrule
\end{tabular}

\vspace{0.2cm}
\footnotesize\textit{Note: PINN and DBAW-PINN do not utilize spline basis functions; thus, the order and grid size parameters are not applicable. The best results are highlighted in bold. }
\end{table*}

\begin{figure}[!t]
  \centering
  \begin{subfigure}[b]{\linewidth}
    \centering
    \includegraphics[width=0.88\linewidth]{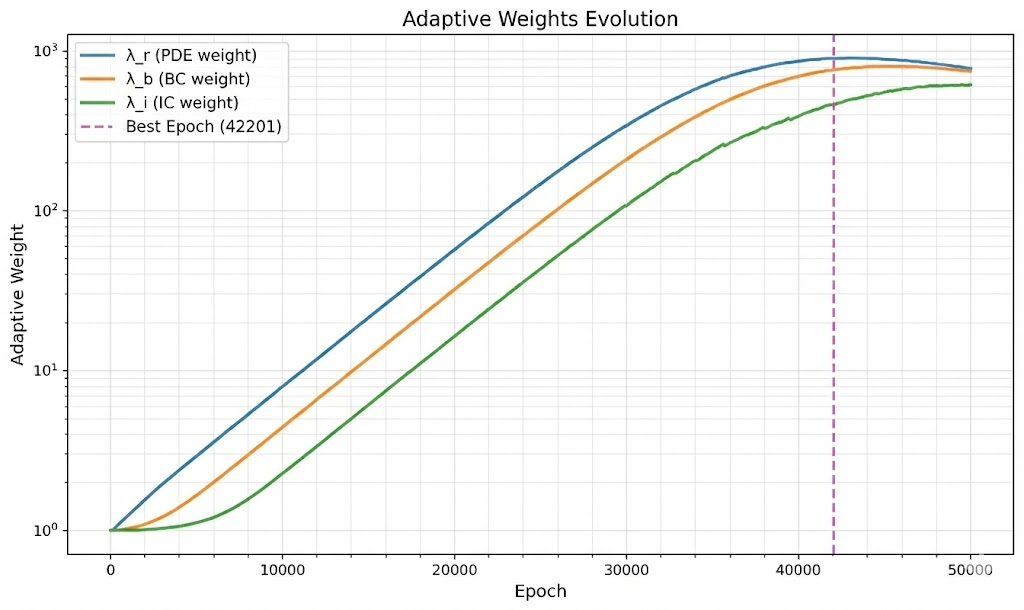}
    \caption{Evolution of adaptive weights}
    \label{fig:BURGERS_weights_evolution}
  \end{subfigure}
  \vspace{0.3em}
  \begin{subfigure}[b]{\linewidth}
    \centering
    \includegraphics[width=0.88\linewidth]{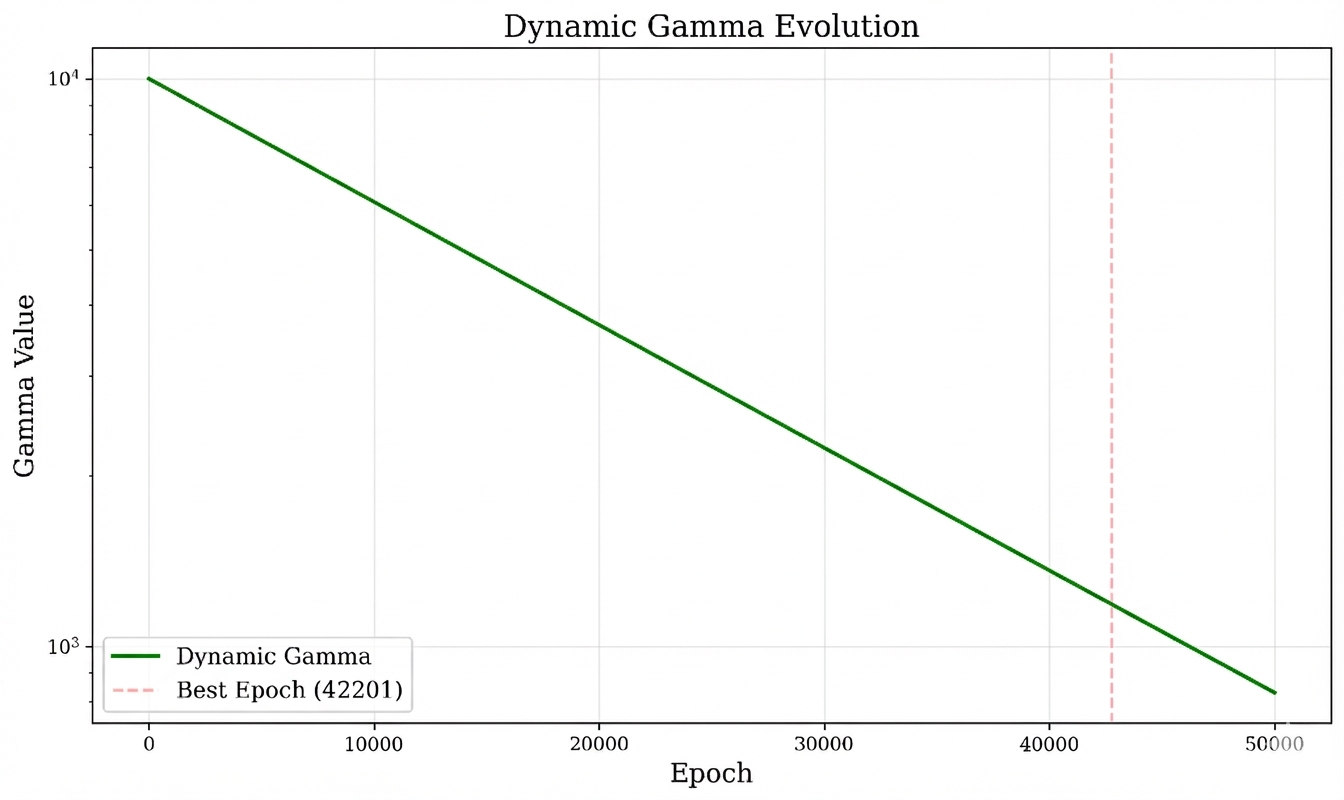}
    \caption{Evolution of the dynamic upper bound $\gamma(t)$}
    \label{fig:burgers_gamma_evolution}
  \end{subfigure}
  \vspace{0.3em}
  \begin{subfigure}[b]{\linewidth}
    \centering
    \includegraphics[width=0.88\linewidth]{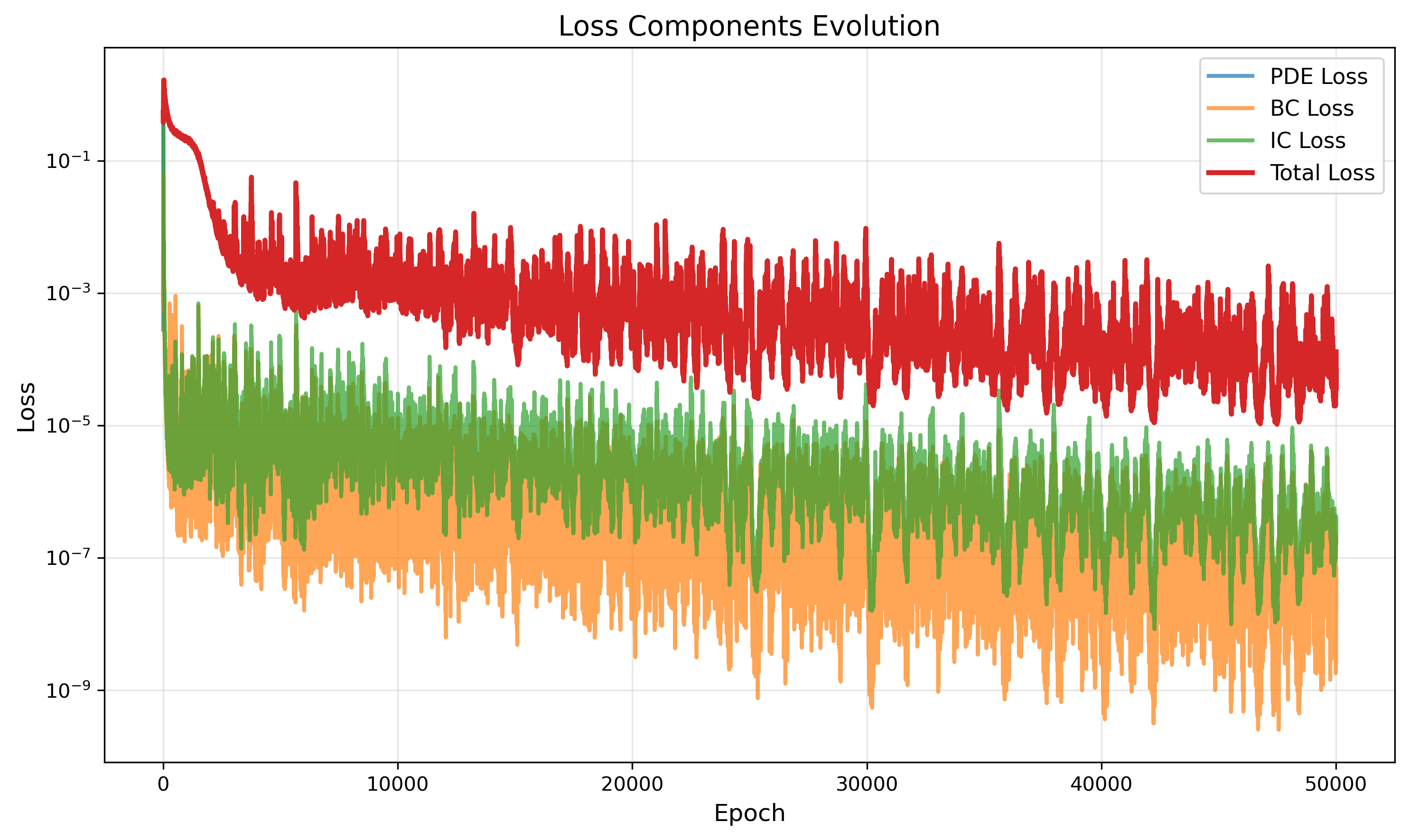}
    \caption{Evolution of loss components}
    \label{fig:BURGERS_loss_components_evolution}
  \end{subfigure}
  \caption{Parameter evolution during the training process for the Burgers equation.}
  \label{fig:BURGERS_evolution}
\end{figure}

\begin{figure*}[htbp]
    \centering

    \begin{minipage}{0.32\textwidth}
        \centering
        \includegraphics[height=4cm, width=\linewidth, keepaspectratio=false]{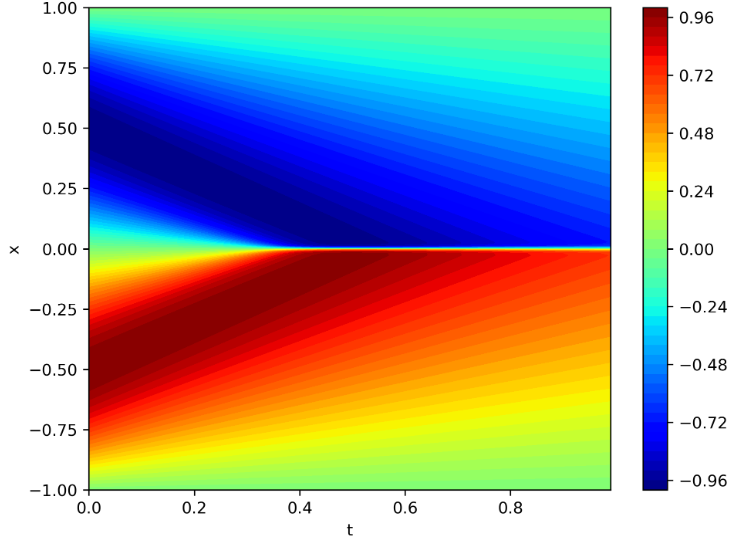}
    \end{minipage}\hfill
    \begin{minipage}{0.32\textwidth}
        \centering
        \includegraphics[height=4cm, width=\linewidth, keepaspectratio=false]{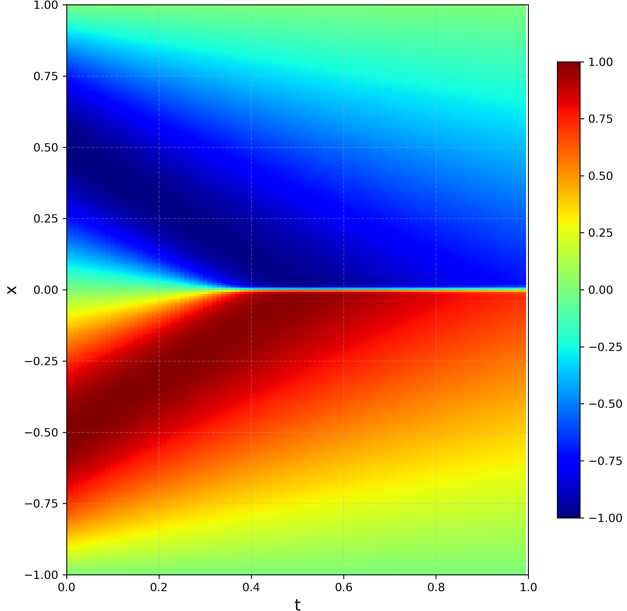}
    \end{minipage}\hfill
    \begin{minipage}{0.32\textwidth}
        \centering
        \includegraphics[height=4cm, width=\linewidth, keepaspectratio=false]{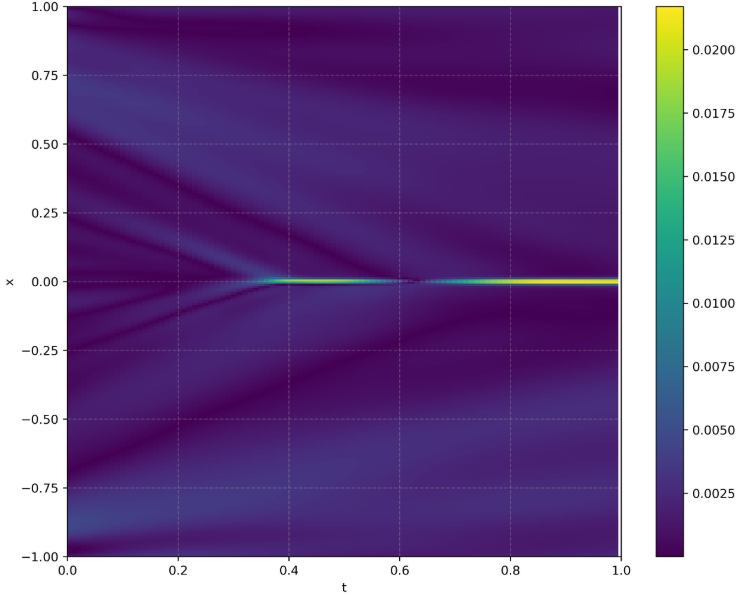}
    \end{minipage}
    \caption*{(a) PINN}

    \vspace{0.3cm}

    \begin{minipage}{0.32\textwidth}
        \centering
        \includegraphics[height=4cm, width=\linewidth, keepaspectratio=false]{Burgers/Burgers_reference.png}
    \end{minipage}\hfill
    \begin{minipage}{0.32\textwidth}
        \centering
        \includegraphics[height=4cm, width=\linewidth, keepaspectratio=false]{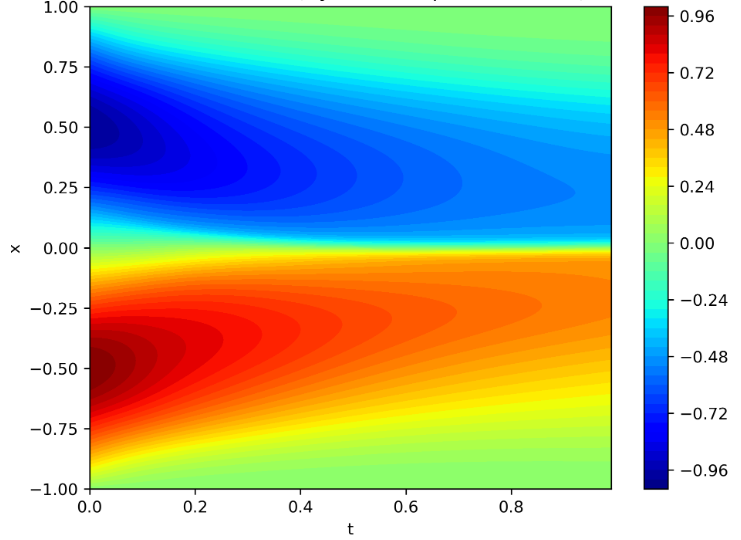}
    \end{minipage}\hfill
    \begin{minipage}{0.32\textwidth}
        \centering
        \includegraphics[height=4cm, width=\linewidth, keepaspectratio=false]{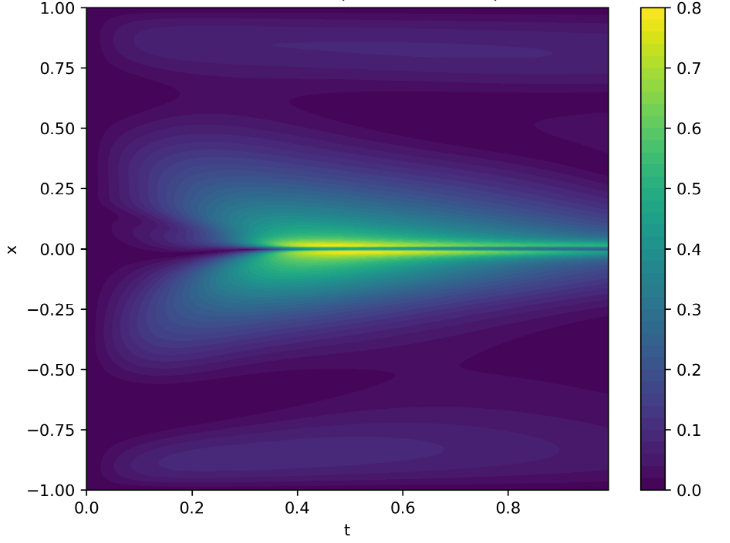}
    \end{minipage}
    \caption*{(b) DBAW-PINN}

    \vspace{0.3cm}

    \begin{minipage}{0.32\textwidth}
        \centering
        \includegraphics[height=4cm, width=\linewidth, keepaspectratio=false]{Burgers/Burgers_reference.png}
    \end{minipage}\hfill
    \begin{minipage}{0.32\textwidth}
        \centering
        \includegraphics[height=4cm, width=\linewidth, keepaspectratio=false]{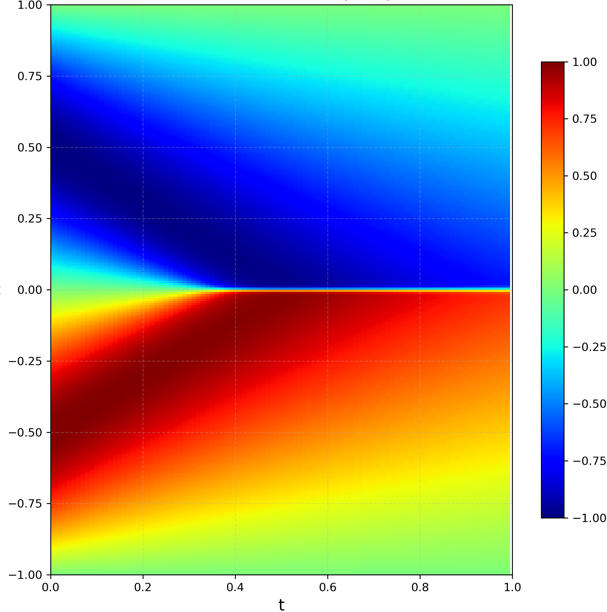}
    \end{minipage}\hfill
    \begin{minipage}{0.32\textwidth}
        \centering
        \includegraphics[height=4cm, width=\linewidth, keepaspectratio=false]{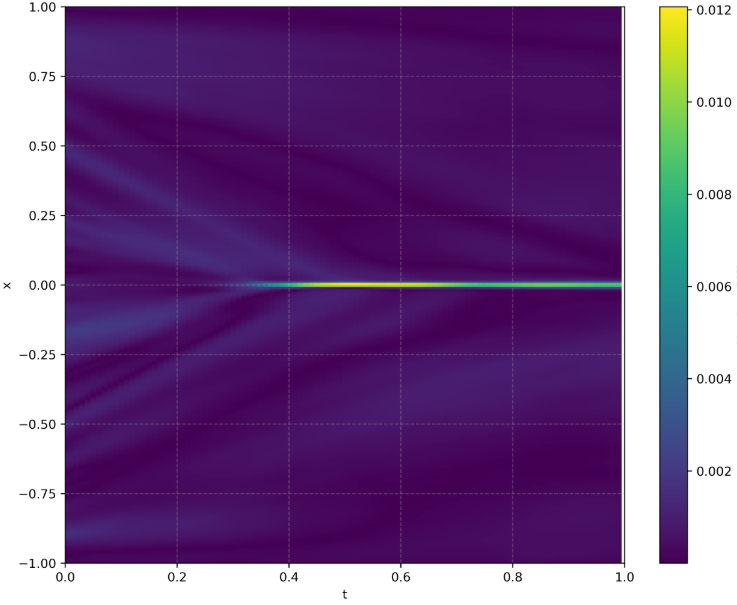}
    \end{minipage}
    \caption*{(c) PIKAN}

    \vspace{0.3cm}

    \begin{minipage}{0.32\textwidth}
        \centering
        \includegraphics[height=4cm, width=\linewidth, keepaspectratio=false]{Burgers/Burgers_reference.png}
    \end{minipage}\hfill
    \begin{minipage}{0.32\textwidth}
        \centering
        \includegraphics[height=4cm, width=\linewidth, keepaspectratio=false]{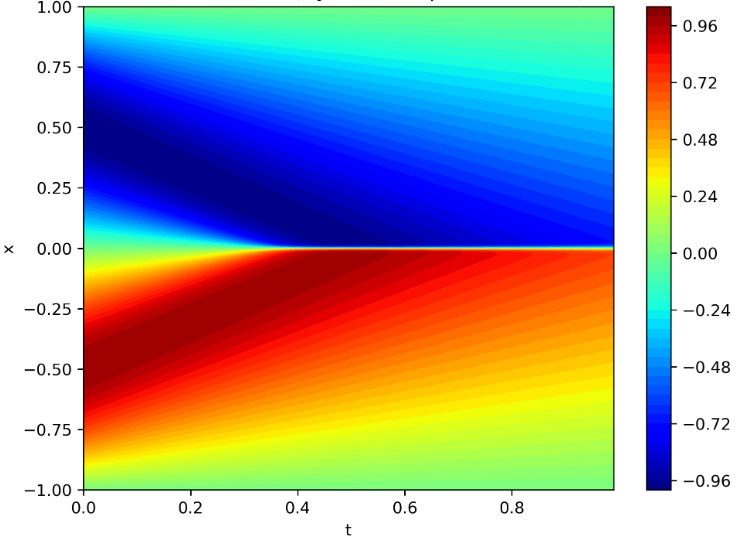}
    \end{minipage}\hfill
    \begin{minipage}{0.32\textwidth}
        \centering
        \includegraphics[height=4cm, width=\linewidth, keepaspectratio=false]{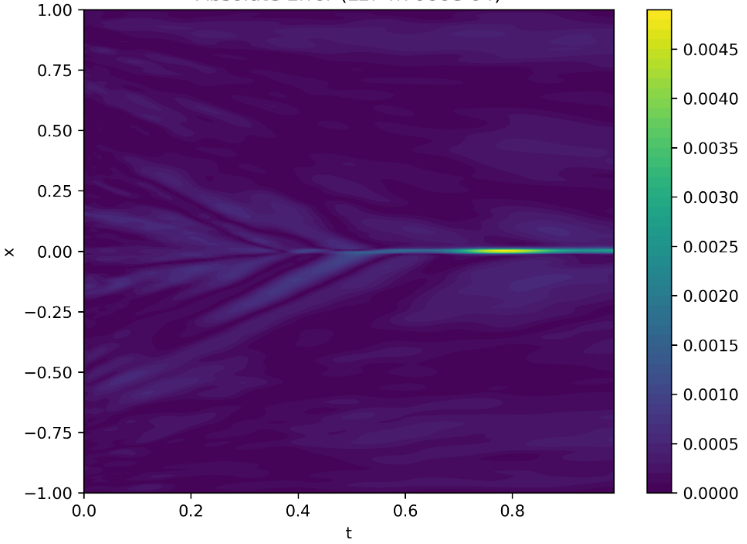}
    \end{minipage}
    \caption*{(d) DBAW-PIKAN}

    \caption{Numerical solution comparisons of the Burgers equation among four different methods. Each row displays (from left to right) the exact solution, the predicted solution, and the point-wise absolute error distribution for each respective framework.}
    \label{fig:burgers_comparison_main}
\end{figure*}

In contrast, the PIKAN model demonstrates superior performance, reducing the error to $1.16 \times 10^{-3}$, which further underscores the potential of KAN in fitting complex nonlinear functions. Ultimately, the proposed DBAW-PIKAN method yields the highest precision with an error of $\mathbf{4.70 \times 10^{-4}}$. This significant improvement over the standard PIKAN validates the effectiveness of integrating the dynamic adaptive weighting strategy with the KAN architecture.
The numerical solutions are visually compared in Fig. \ref{fig:burgers_comparison_main}: (a) The PINN prediction shows visible deviations from the reference solution near the shock wave region. (b) The DBAW-PINN prediction severely deviates from the reference. (c) The PIKAN prediction is in good overall agreement with the reference. (d) The DBAW-PIKAN solution is visually the most accurate, particularly in precisely capturing the shock front, with its corresponding error map exhibiting the lowest overall error level.
To further elucidate the dynamic optimization mechanism of DBAW-PIKAN when solving the nonlinear Burgers equation, Fig. \ref{fig:BURGERS_evolution} illustrates the evolution trajectories of key parameters during training. As shown in Fig. \ref{fig:BURGERS_weights_evolution}, the boundary condition weight $\lambda_{bc}$ (orange curve) escalates rapidly during the initial phase. This aligns with the characteristic training behavior of PINNs, where simple functional relationships on the boundaries are typically easier for the network to fit than the complex interior PDE residuals. Without intervention, such a rapid increase in weight would cause the optimizer to over-focus on the boundary loss, thereby neglecting the learning of the underlying PDE physics.

However, the regulatory role of the DBAW strategy is clearly visible from the comparison between Fig. \ref{fig:BURGERS_weights_evolution} and Fig. \ref{fig:burgers_gamma_evolution}. As the dynamic upper bound parameter $\gamma(t)$ decays exponentially, the growth of $\lambda_{bc}$ is forcibly clamped and gradually recedes in the middle stage of training. Meanwhile, $\lambda_{r}$ (blue curve), representing the importance of PDE residuals, maintains a relatively high preponderance throughout the later stages. This mandatory balancing mechanism is corroborated by the evolution of loss components in Fig.\ref{fig:BURGERS_loss_components_evolution}. Although $\mathcal{L}_{BC}$ (green curve) drops sharply in the early stages, its contribution to the total gradient is confined within a reasonable range under the constraint of $\gamma(t)$, allowing the model to consistently focus on minimizing the PDE residual loss $\mathcal{L}_{PDE}$ (red curve). This weight balancing, achieved through dynamic constraints, effectively circumvents the training instability caused by gradient flow pathologies, ensuring a smooth optimization process. Consequently, the model converges to a high-precision level of the order of $10^{-4}$ as shown in Table II, fully demonstrating the effectiveness of the strategy in handling strongly nonlinear fluid dynamics problems.
Overall, while DBAW-PINN performs suboptimally in this case, the success of DBAW-PIKAN further highlights the importance of combining an advanced network architecture (KAN) with a compatible optimization strategy (DBAW). The experimental results indicate that the DBAW-PIKAN framework possesses high accuracy and robustness for solving nonlinear advection-diffusion problems like the Burgers equation.

\subsection{The Helmholtz Equation}
The Helmholtz equation is a fundamental partial differential equation in mathematical physics, appearing in various fields such as acoustics, electromagnetics, and quantum mechanics. It describes the behavior of wave phenomena, such as the propagation of acoustic and electromagnetic waves within a specific domain. This equation is frequently used to model problems under steady-state conditions, particularly when analyzing wave behavior in confined regions. The general form of the Helmholtz equation is expressed as:
\begin{equation}
\nabla^2 u + k^2 u = q(x,y), \quad (x,y) \in \Omega = [-1,1]^2,
\end{equation}
where $\nabla^2$ is the Laplace operator, $u=u(x,y)$ is the unknown wave field function, $k$ is the wave number that controls the wave propagation characteristics, and $q(x,y)$ is the forcing term representing external excitation sources, defined as:
\begin{equation}
q(x,y) = -(a_1\pi)^2 u - (a_2\pi)^2 u + k^2 u.
\end{equation}
Following \cite{raissi2019physics}, the analytical solution for this equation is given by:
\begin{equation}
u(x,y) = \sin(a_1\pi x) \sin(a_2\pi y).
\end{equation}
This analytical solution is employed to validate the model performance, where $a_1$ and $a_2$ govern the frequency of the spatial distribution. For this problem, the boundary conditions are specified as:
\begin{equation}
u(-1, y) = u(1, y) = u(x, -1) = u(x, 1) = 0.
\end{equation}

In this experiment, the parameters are set to $k = 1, a_1 = 1$, and $a_2 = 4$. The training dataset consists of $N_r = 5000$ interior collocation points and $N_{bc} = 400$ boundary condition points. Boundary points are sampled using a uniform distribution to ensure sufficient coverage along the domain boundaries. All models are trained for 50,000 iterations, and the results are summarized in Table \ref{tab:helmholtz_comparison}.Table \ref{tab:helmholtz_comparison} details the performance metrics of the four methods for solving the Helmholtz equation. Quantitatively, the DBAW-PIKAN model achieves the lowest relative $L_2$ error of $\mathbf{4.17 \times 10^{-4}}$. In comparison, the baseline PINN model yields the highest error at $1.82 \times 10^{-2}$. The introduction of dynamic adaptive weighting to the PINN framework (DBAW-PINN) reduces the error to $1.08 \times 10^{-2}$, initially validating the effectiveness of the adaptive weighting strategy. Furthermore, replacing the MLP architecture with KAN (PIKAN) significantly reduces the error to $1.32 \times 10^{-3}$, showcasing the advantages of KAN in function approximation. Ultimately, the DBAW-PIKAN model, which integrates both enhancement strategies, improves the accuracy by nearly an order of magnitude compared to PIKAN, demonstrating the effectiveness of the integrated framework.

To visually assess the model performance, Fig. \ref{fig:helmholtz_results1} presents the predicted results of the four methods for the Helmholtz equation. Each row corresponds to a specific method, displaying the exact solution, the predicted solution, and the point-wise absolute error distribution from left to right:(a) PINN method: The predicted solution shows obvious visual discrepancies compared to the exact solution. The extensive bright regions in the absolute error plot indicate significant errors across most of the domain.(b) DBAW-PINN method: The prediction results are improved compared to standard PINN, and the overall intensity of the error map is reduced, though the error distribution remains relatively broad.(c) PIKAN method: The predicted solution is more morphologically consistent with the exact solution, and the significantly darker color in the error map indicates a substantial reduction in error magnitude.(d) DBAW-PIKAN method: Our proposed method yields a predicted solution that is in high visual agreement with the exact solution. Its absolute error plot features the deepest color tones, with errors concentrated in localized regions of wave peaks and troughs, demonstrating the highest solution precision.

\begin{figure}[htbp]
  \centering
  \begin{subfigure}[b]{\linewidth}
    \centering
    \includegraphics[width=\linewidth]{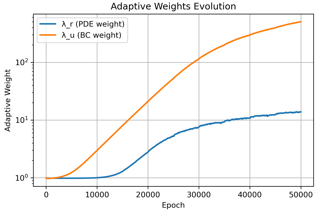}
    \caption{Evolution of adaptive weights}
    \label{fig:helmhotz_weights_evolution} 
  \end{subfigure}
  \hfill
  \begin{subfigure}[b]{\linewidth}
    \centering
    \includegraphics[width=\linewidth]{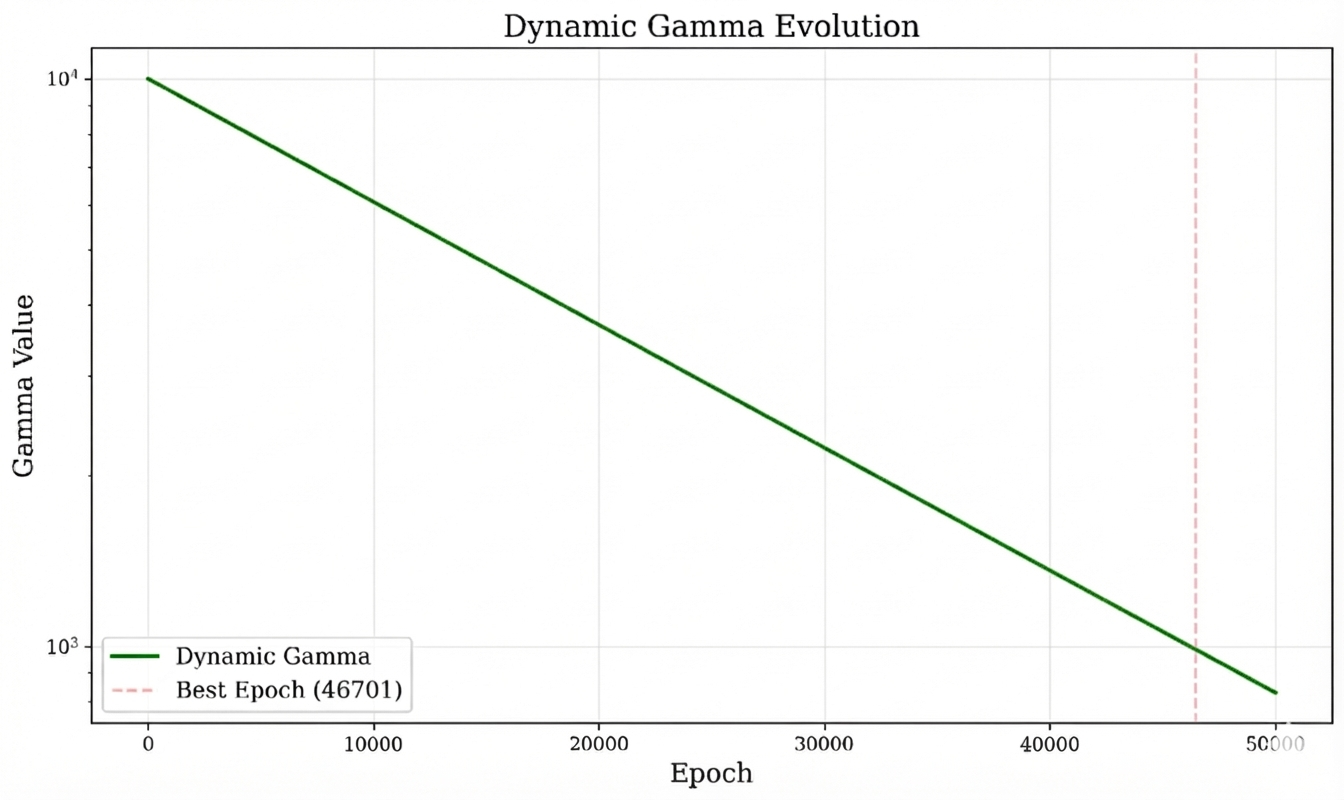}
    \caption{Evolution of the dynamic upper bound}
    \label{fig:helmhotz_gamma_evolution}
  \end{subfigure}
  \hfill
  \begin{subfigure}[b]{\linewidth}
    \centering
    \includegraphics[width=\linewidth]{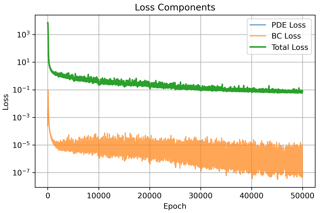}
    \caption{Evolution of loss components}
    \label{fig:helmhotz_loss_components_evolution}
  \end{subfigure}
  
  \caption{Parameter evolution during the training process for the Helmholtz equation: (a) Evolution of adaptive weights ($\lambda_r, \lambda_{ic}, \lambda_{bc}$) over training epochs; (b) Evolution of the dynamic upper bound parameter $\gamma(t)$ over training epochs; (c) Evolution of individual loss components over training epochs.}
  \label{fig:helmhotz_evolution}
\end{figure}

\begin{figure*}[htbp]
    \centering
    \begin{subfigure}{\textwidth}
        \centering
        \begin{subfigure}{0.3\textwidth}
            \centering
            \includegraphics[width=\textwidth]{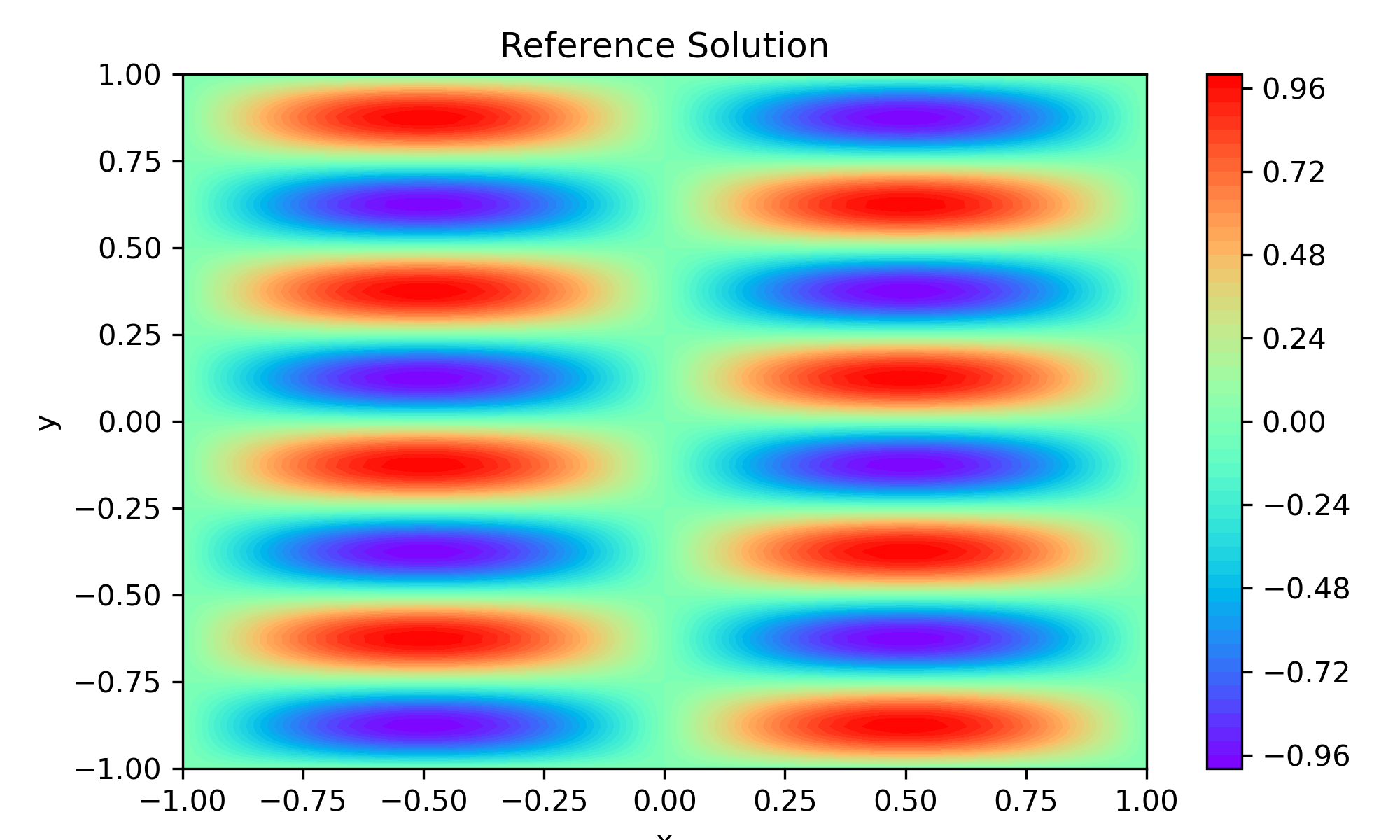}
            \caption{}
            \label{fig:Helmholtz_mlp_pikan_exact}
        \end{subfigure}
        \hfill
        \begin{subfigure}{0.3\textwidth}
            \centering
            \includegraphics[width=\textwidth]{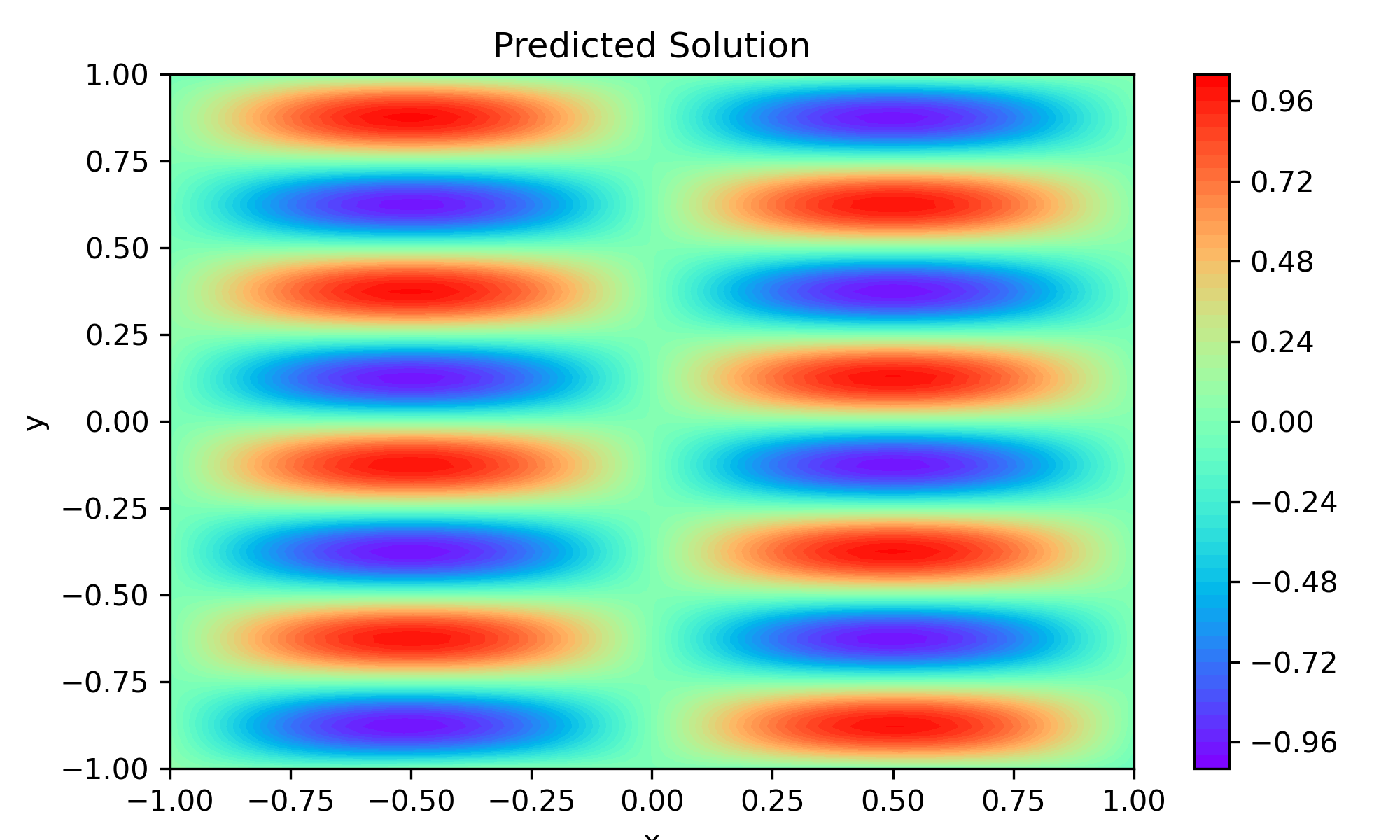}
            \caption{}
            \label{fig:Helmholtz_mlp_pikan_predicted}
        \end{subfigure}
        \hfill
        \begin{subfigure}{0.3\textwidth}
            \centering
            \includegraphics[width=\textwidth]{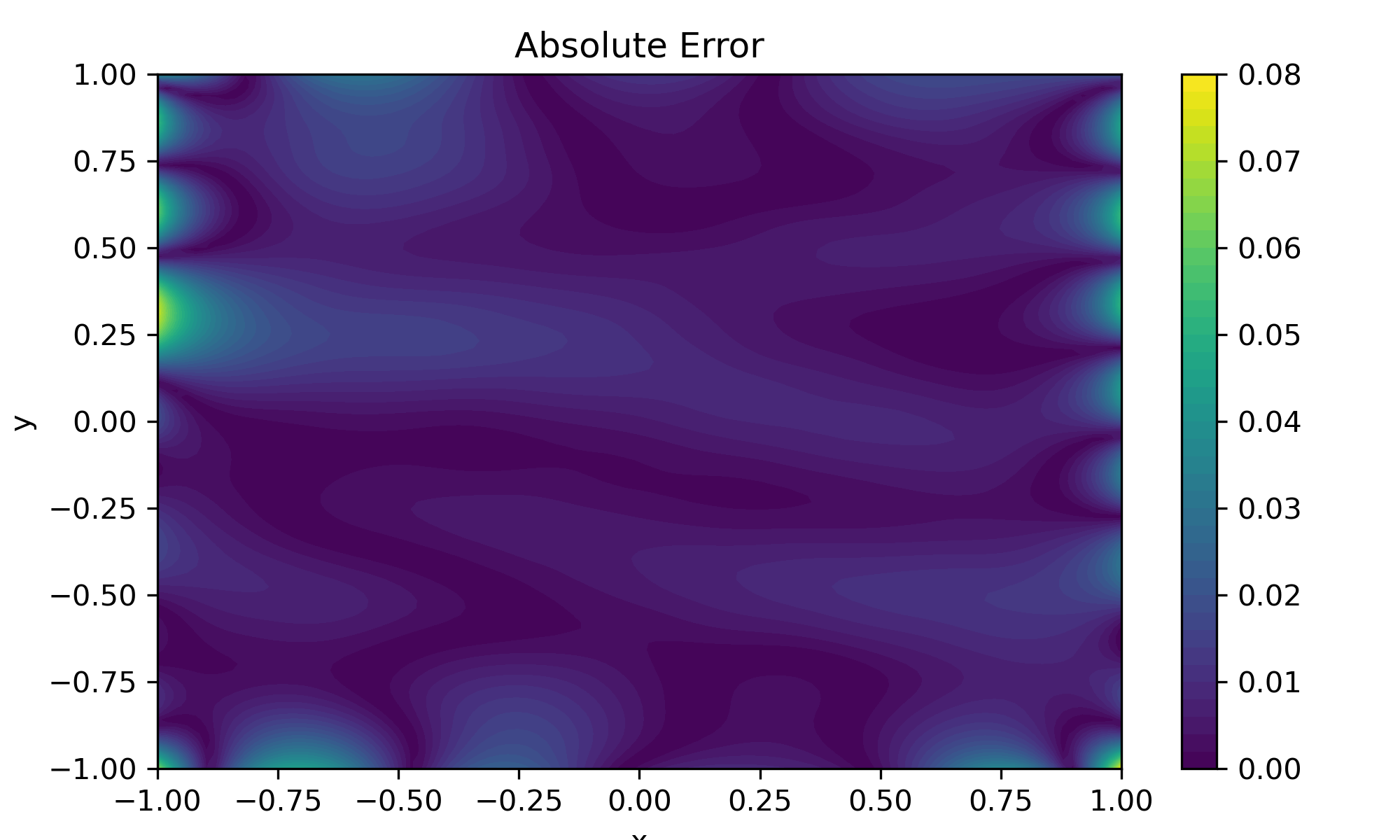}
            \caption{}
            \label{fig:Helmholtz_mlp_pikan_error}
        \end{subfigure}
        \caption*{(a) PINN}
    \end{subfigure}
    
    \vspace{0.5cm} 
    \begin{subfigure}{\textwidth}
        \centering
        \begin{subfigure}{0.3\textwidth}
            \centering
            \includegraphics[width=\textwidth]{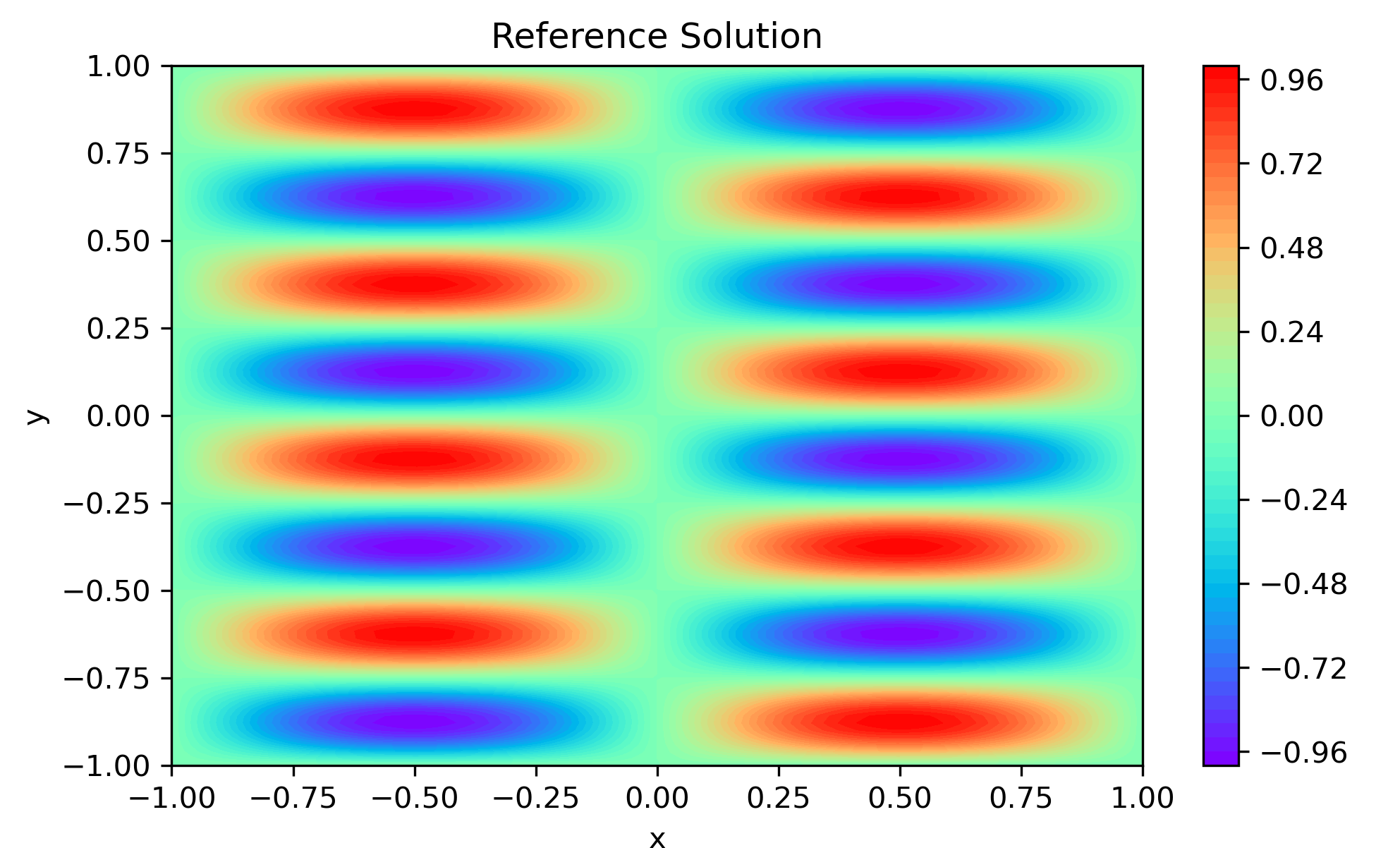}
            \caption{}
            \label{fig:Helmholtz_DBAWMLP_pikan_exact}
        \end{subfigure}
        \hfill
        \begin{subfigure}{0.3\textwidth}
            \centering
            \includegraphics[width=\textwidth]{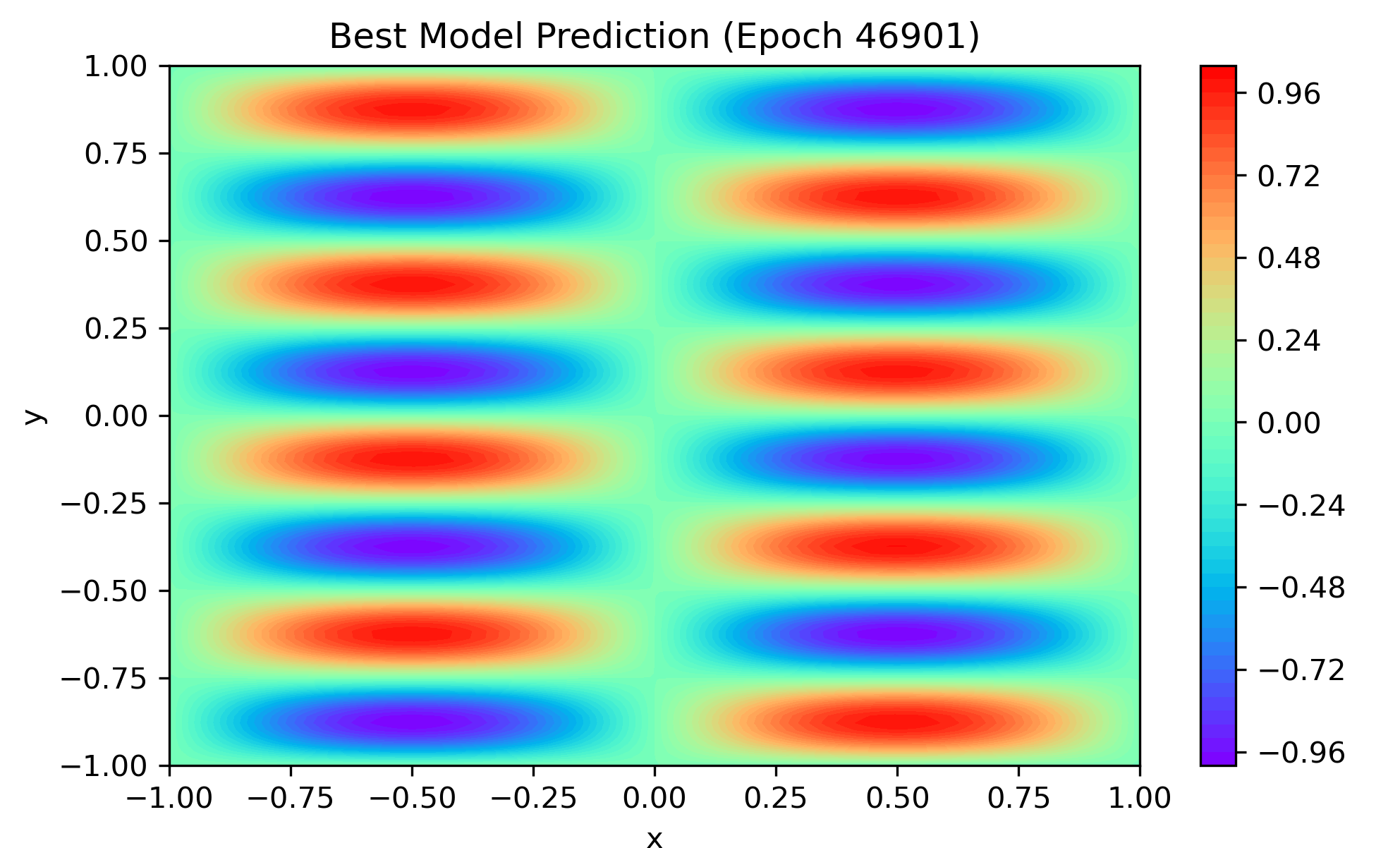}
            \caption{}
            \label{fig:Helmholtz_DBAWMLP_pikan_predicted}
        \end{subfigure}
        \hfill
        \begin{subfigure}{0.3\textwidth}
            \centering
            \includegraphics[width=\textwidth]{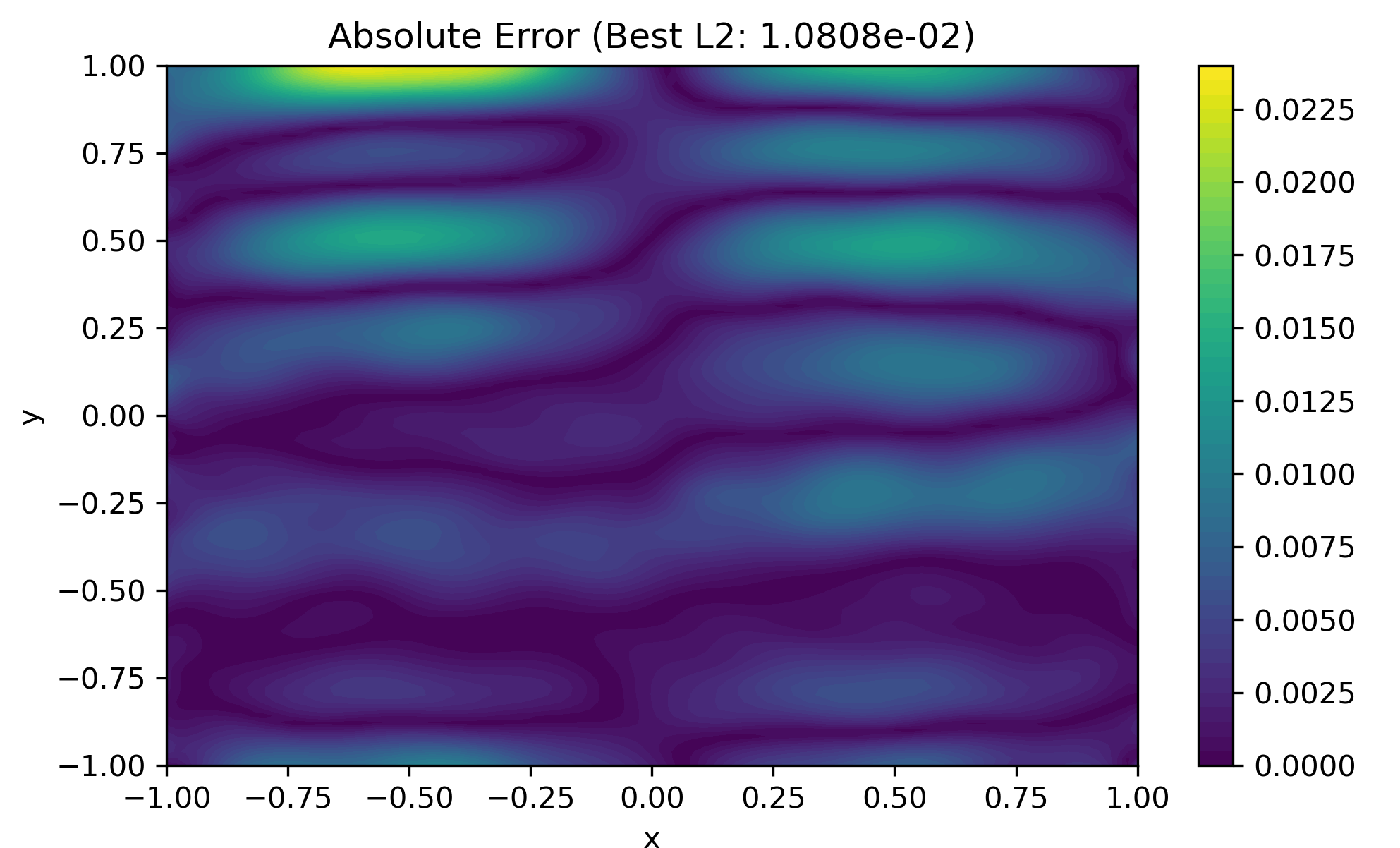}
            \caption{}
            \label{fig:Helmholtz_DBAWMLP_pikan_error}
        \end{subfigure}
        \caption*{(b) DBAW-PINN}
    \end{subfigure}
    
    \vspace{0.5cm} 
    \begin{subfigure}{\textwidth}
        \centering
        \begin{subfigure}{0.3\textwidth}
            \centering
            \includegraphics[width=\textwidth]{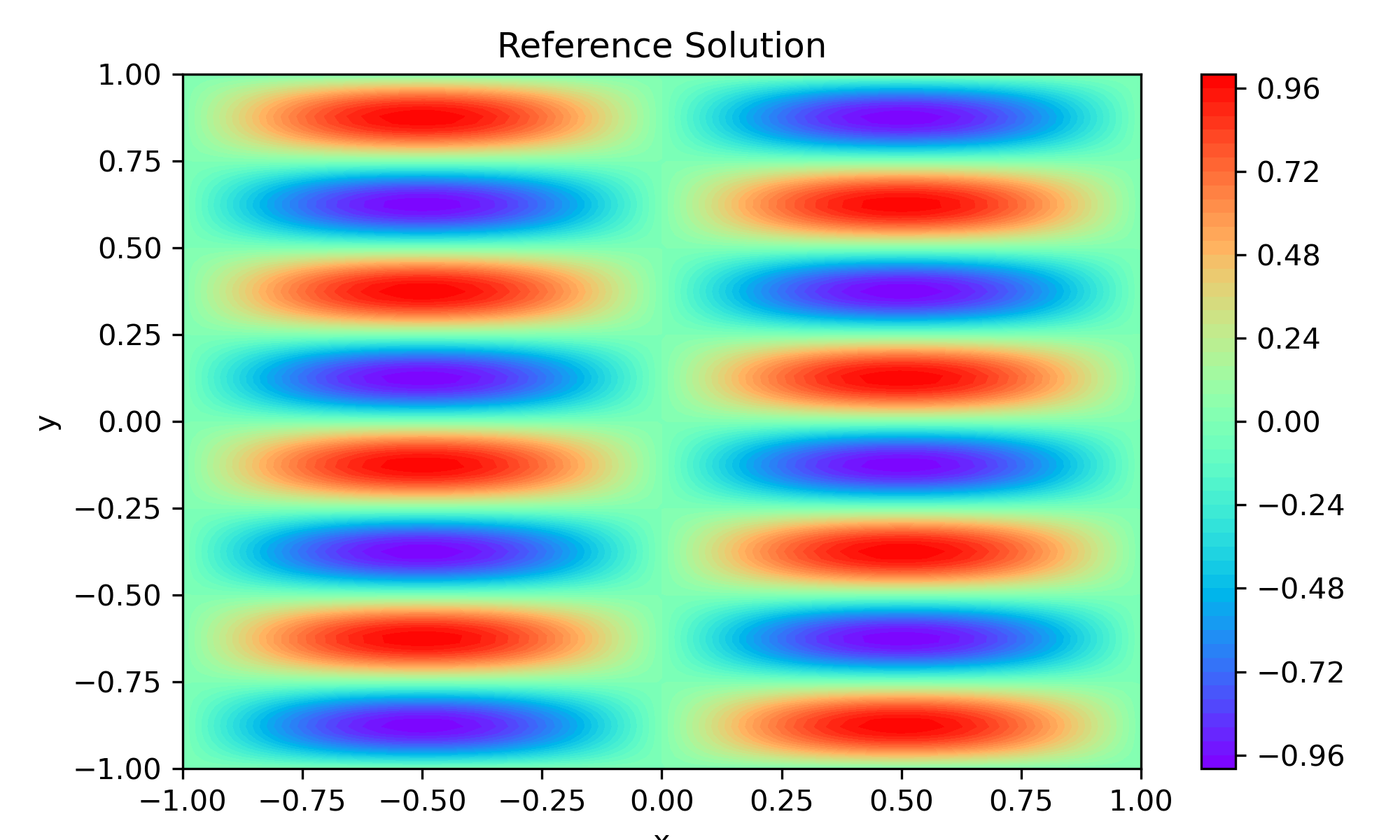}
            \caption{}
            \label{fig:Helmholtz_pikan_exact}
        \end{subfigure}
        \hfill
        \begin{subfigure}{0.3\textwidth}
            \centering
            \includegraphics[width=\textwidth]{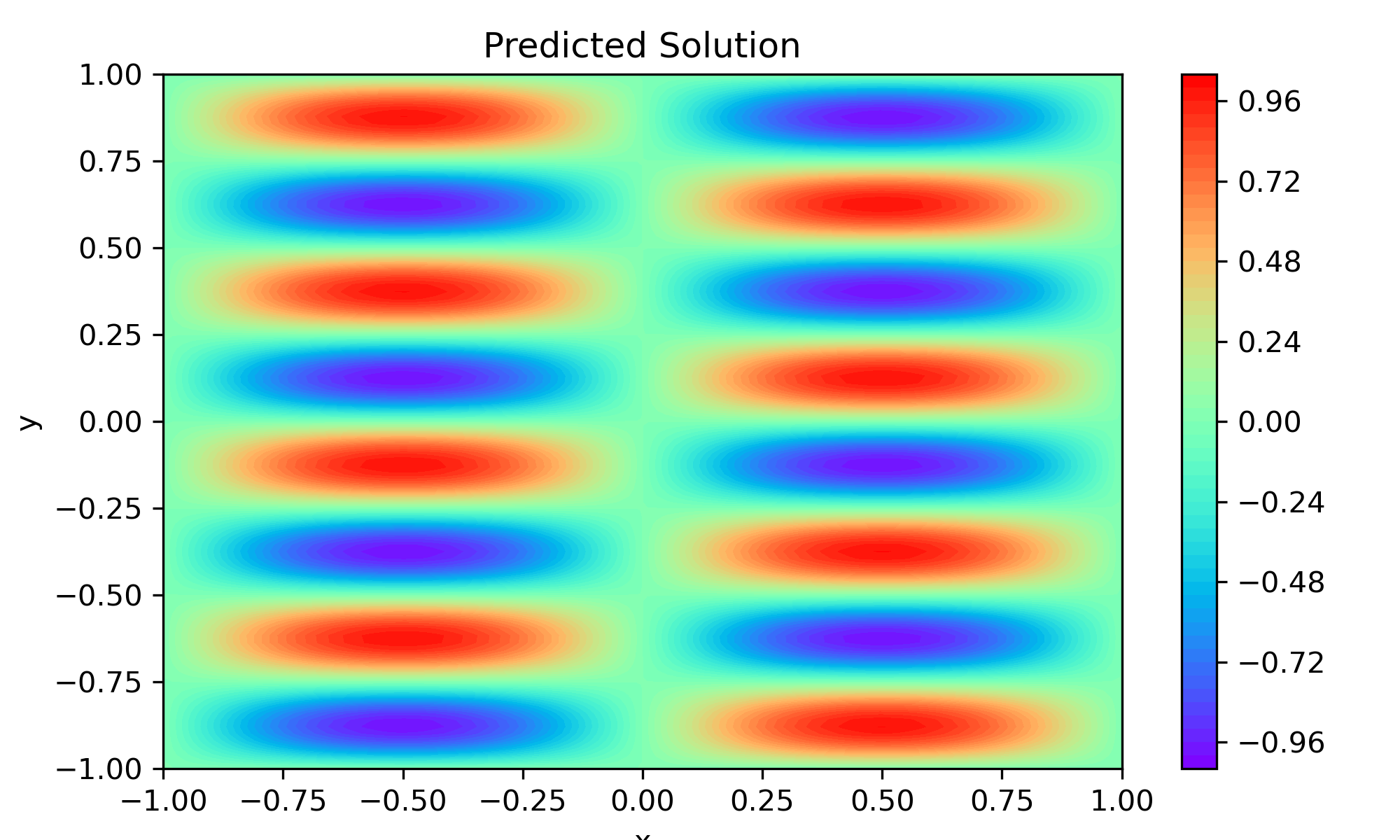}
            \caption{}
            \label{fig:pikan_predicted}
        \end{subfigure}
        \hfill
        \begin{subfigure}{0.3\textwidth}
            \centering
            \includegraphics[width=\textwidth]{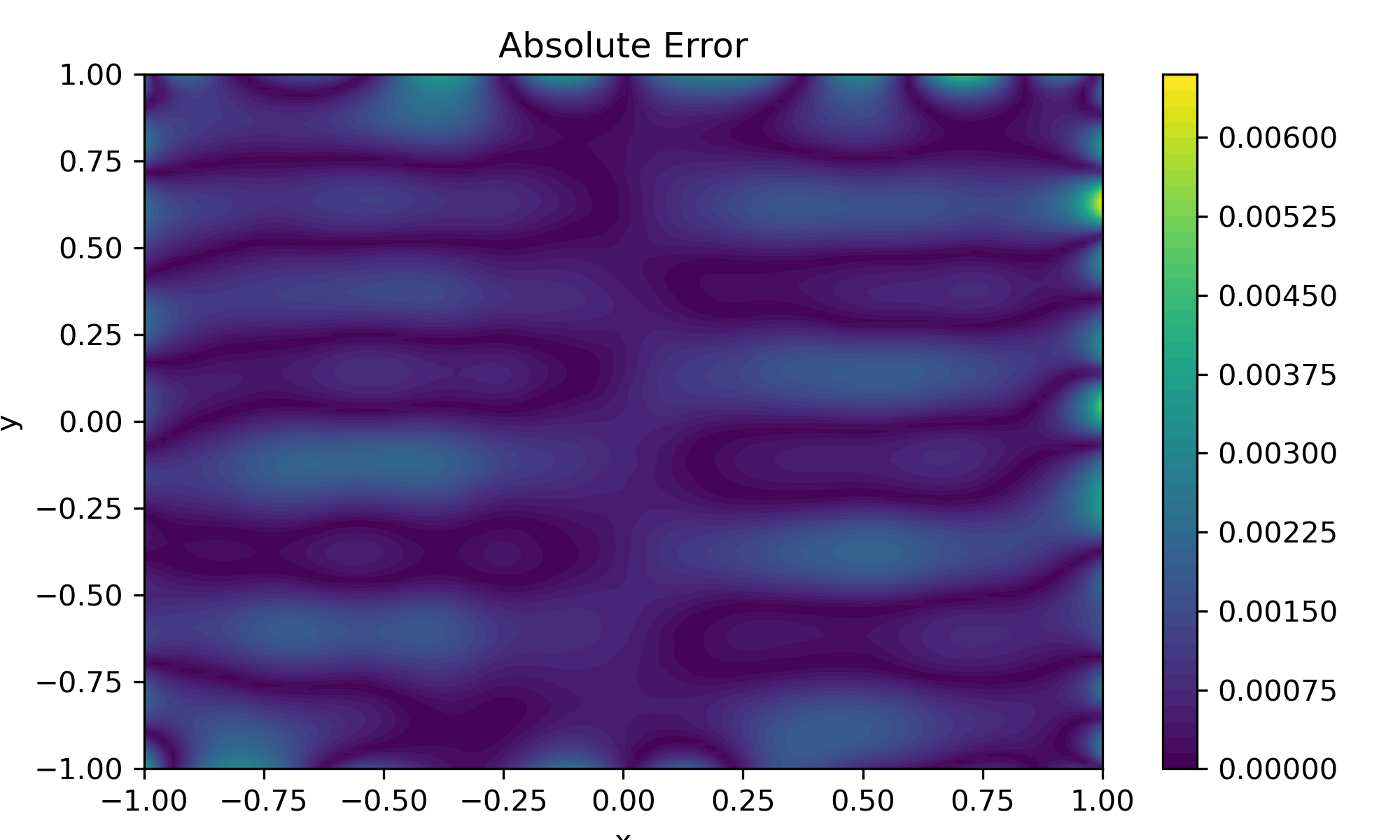}
            \caption{}
            \label{fig:Helmholtz_pikan_error}
        \end{subfigure}
        \caption*{(c) PIKAN}
    \end{subfigure}
    
    \vspace{0.5cm} 
    
    \begin{subfigure}{\textwidth}
        \centering
        \begin{subfigure}{0.3\textwidth}
            \centering
            \includegraphics[width=\textwidth]{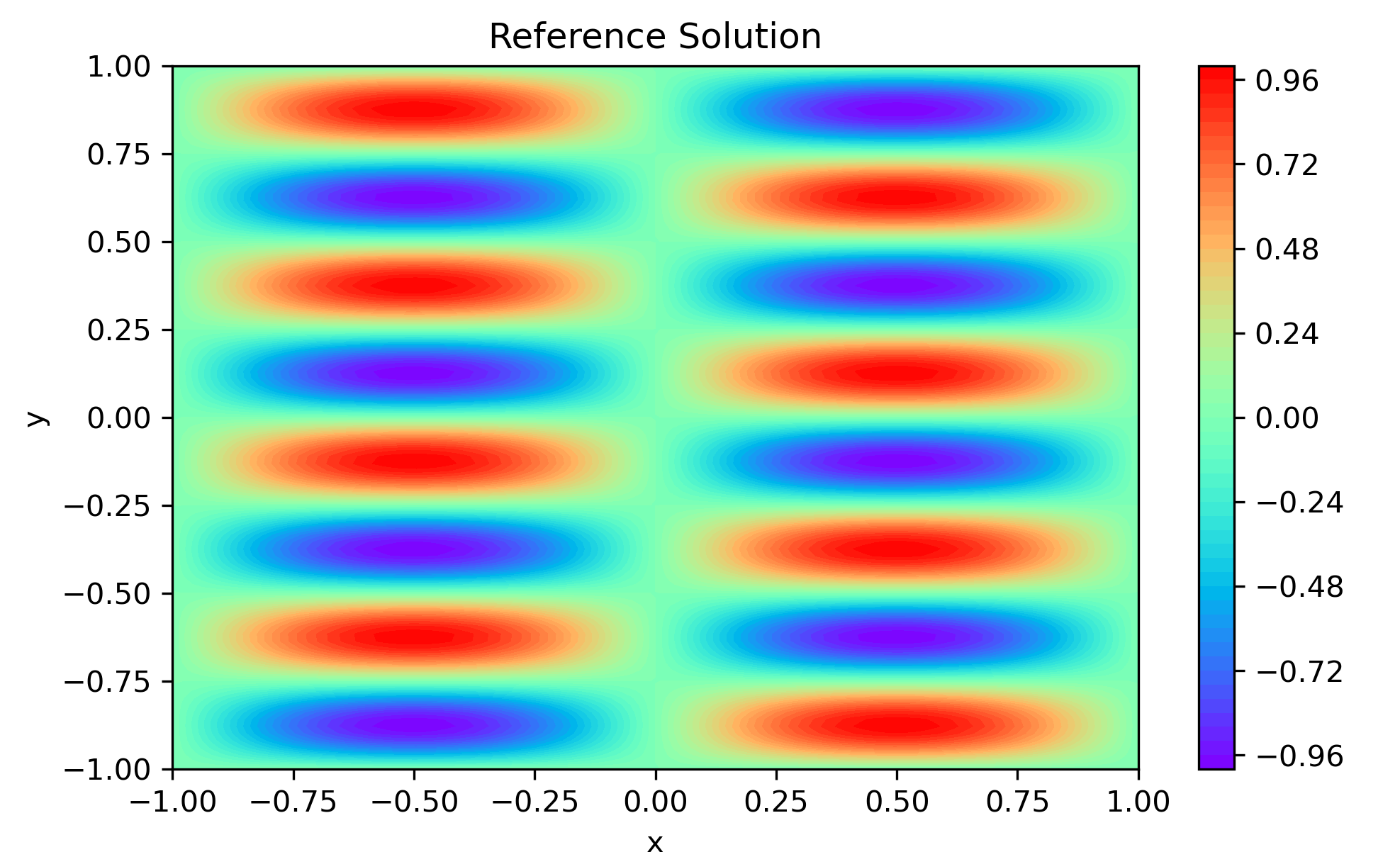}
            \caption{}
            \label{fig:Helmholtz_dbaw_exact}
        \end{subfigure}
        \hfill
        \begin{subfigure}{0.3\textwidth}
            \centering
            \includegraphics[width=\textwidth]{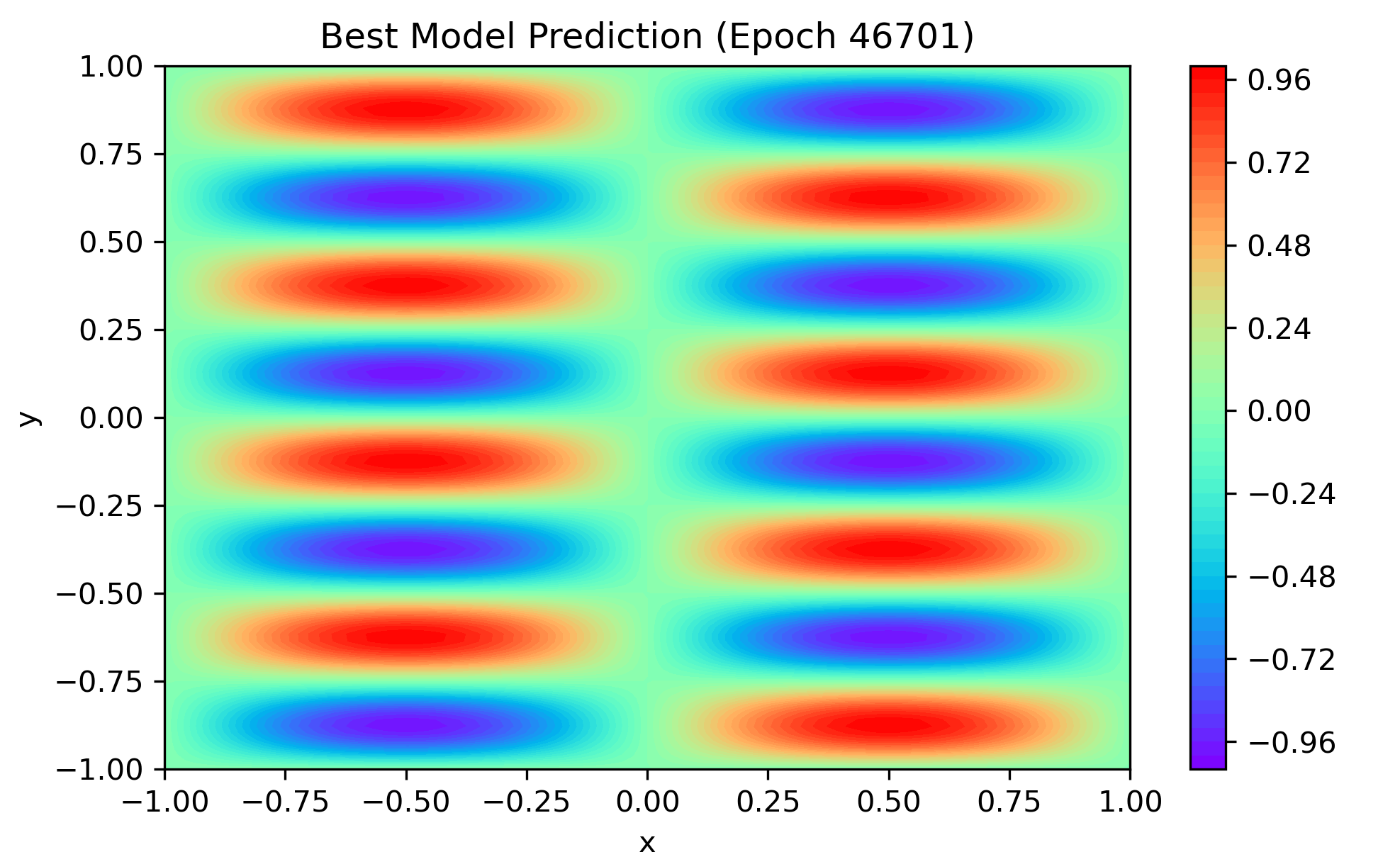}
            \caption{}
            \label{fig:Helmholtz_dbaw_predicted}
        \end{subfigure}
        \hfill
        \begin{subfigure}{0.3\textwidth}
            \centering
            \includegraphics[width=\textwidth]{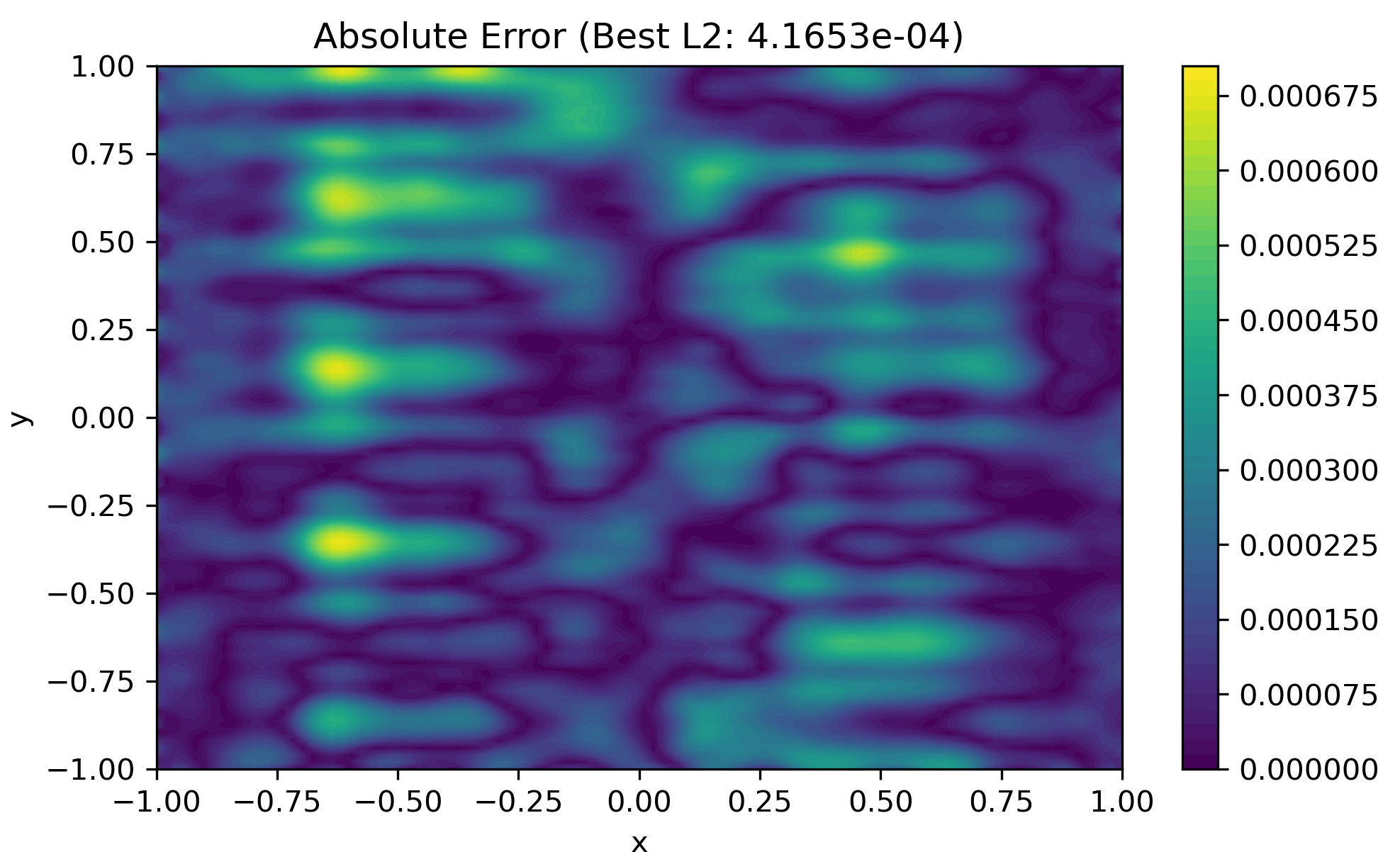}
            \caption{}
            \label{fig:Helmholtz_dbaw_error}
        \end{subfigure}
        \caption*{(d) DBAW-PIKAN}
    \end{subfigure}
    
    \caption{Numerical solution comparisons for the Helmholtz equation among four different methods: (a) PINN, (b) DBAW-PINN, (c) PIKAN, and (d) DBAW-PIKAN. Each row displays the exact solution, predicted solution, and point-wise absolute error from left to right.}
    \label{fig:helmholtz_results1}
    
\end{figure*}

To further reveal the dynamic optimization mechanism of DBAW-PIKAN for the nonlinear Helmholtz equation, Fig. \ref{fig:helmhotz_evolution} illustrates the evolution trajectories of key parameters during training. The solution to the Helmholtz equation possesses pronounced spatial oscillatory characteristics, making the fitting of PDE residuals highly challenging. As shown in Fig. 8c, the magnitude of the PDE residual loss $\mathcal{L}_{PDE}$ remains consistently higher than the boundary loss $\mathcal{L}_{BC}$. In traditional adaptive weighting methods, such a large magnitude discrepancy often leads to excessively small PDE weights, resulting in "high-frequency distortion."

However, under the regulation of the DBAW strategy, Fig. 8a shows that while the boundary weight $\lambda_{bc}$ initially attempts to grow rapidly, it is immediately suppressed by the dynamic upper bound $\gamma(t)$ (Fig. 8b). More importantly, this strategy ensures that the PDE residual weight $\lambda_r$ remains at a similar order of magnitude as the boundary weight throughout the training cycle, even becoming slightly dominant in the later stages. This dynamic "rectification" of weights forces the KAN network to maintain sufficient focus on the governing PDE during optimization. Thanks to this mechanism, the model avoids falling into local minima while handling the complex peak-and-trough structures of the Helmholtz equation. The steady decline of the total loss $\mathcal{L}_{Total}$ in Fig. 8c, combined with the extremely low error of $4.17 \times 10^{-4}$ in Table III, proves that the DBAW strategy effectively coordinates the learning progress of different frequency features for multi-scale oscillatory problems, thereby fully unlocking the potential of the KAN architecture in function approximation.

In summary, both quantitative $L_2$ errors and qualitative visual error distributions consistently indicate that the DBAW-PIKAN framework provides a significant performance advantage when solving wave problems such as the Helmholtz equation compared to baseline models.

\begin{table*}[htbp]
\centering
\caption{Performance comparison of different methods for solving the Helmholtz equation.}
\label{tab:helmholtz_comparison}
\small
\begin{tabular}{lccccccc}
\toprule
\textbf{Method} & \textbf{Order} & \textbf{Grid Size} & \textbf{Architecture} & \textbf{Parameters} & \textbf{Optimizer} & \textbf{Iterations} & \textbf{Relative $L_2$ Error} \\
\midrule
PINN & -- & -- & [2,64,64,64,64,64,64,1] & 21,057 & Adam & 50,000 & $1.82 \times 10^{-2}$ \\
DBAW-PINN & -- & -- & [2,64,64,64,64,64,64,1] & 21,057 & Adam & 50,000 & $1.08 \times 10^{-2}$ \\
PIKAN & 4 & 20 & [2,20,20,20,1] & 22,360 & Adam & 50,000 & $1.32 \times 10^{-3}$ \\
\textbf{DBAW-PIKAN} & 4 & 20 & [2,20,20,20,1] & 22,360 & Adam & 50,000 & $\mathbf{4.17 \times 10^{-4}}$ \\ 
\bottomrule
\end{tabular}

\vspace{0.2cm}
\footnotesize\textit{Note: PINN and DBAW-PINN do not utilize spline basis functions; thus, the order and grid size parameters are not applicable. The best results are highlighted in bold.}
\end{table*}

\section{Conclusion}
Addressing the dual challenges of architectural expressivity bottlenecks and multi-objective optimization imbalance encountered by PINNs in solving PDEs, we present an integrated framework termed DBAW-PIKAN . This framework synergistically combines the superior function approximation capabilities of KAN with a novel DBAW strategy characterized by a dynamically decaying upper bound constraint.

The primary contributions of this research are summarized in the following three aspects:
\begin{itemize}
\item Architectural Innovation: By replacing the traditional MLP with KAN as the backbone architecture, DBAW-PIKAN effectively captures the intricate features of PDE solutions and significantly mitigates the spectral bias issue.
\item Refinement of Optimization Strategy: The introduced DBAW strategy utilizes a dynamic decay parameter $\gamma(t)$ to constrain the adaptive weights. This mechanism effectively alleviates the optimization instability caused by the excessive growth of weights in the early stages of training, ensuring that loss terms of different magnitudes receive balanced and continuous optimization.
\item Performance Validation: Systematic numerical experiments on several benchmark PDEs—including the Helmholtz, Burgers, and nonlinear Klein-Gordon equations—demonstrate the efficacy of the DBAW-PIKAN framework. The results show that compared to baseline models such as standard PINN, DBAW-PINN, and PIKAN, the proposed method achieves consistent and significant improvements in accuracy, in some cases by nearly an order of magnitude. Furthermore, the mechanistic analysis of weight and loss dynamics elucidates the critical role of the DBAW strategy in achieving stable and balanced optimization.
\end{itemize}

Despite the promising results, we open several avenues for future exploration. First, while the DBAW strategy proved robust in our experiments, it introduces hyperparameters such as $\gamma_{\max}$, $\gamma_{\min}$, and $\alpha$, the optimal selection of which currently relies on empirical tuning. Future research could investigate automated hyperparameter optimization or explore self-adaptive constraint mechanisms that eliminate the need for manual decay rate settings. Second, this work primarily validates the model on (1+1)-dimensional problems; its scalability and computational efficiency for high-dimensional PDEs warrant further evaluation. Finally, applying the DBAW-PIKAN framework to more complex real-world engineering scenarios—such as turbulence simulation, multi-physics coupling, or parameter inversion from experimental data—will be a crucial next step in verifying its practical utility in science and engineering. In summary, the DBAW-PIKAN framework provides an effective solution to the core challenges facing PINNs, demonstrating the significant potential of combining advanced network architectures with carefully designed optimization strategies and offering a new perspective for the development of scientific machine learning.

\section*{Acknowledgements}
This work is supported by the Scientific Research Foundation of Fujian University of Technology, China (Grant No. GY-Z24010). The authors also thank the anonymous reviewers for their constructive comments that helped improve this paper.

\bibliography{myReference}
\bibliographystyle{elsarticle-num}
\end{document}